\documentclass{arxiv}

\title{Coarsening Causal DAG Models}

\usepackage[american]{babel}
\usepackage{mathtools}
\usepackage{extarrows}
\usepackage{mathbbol}
\usepackage{float}            
\usepackage[nameinlink]{cleveref}
\usepackage{times}

\usepackage{standalone}
\usepackage{graphicx}
\graphicspath{{figures/}}
\usepackage{wrapfig}
\usepackage{caption}
\usepackage{booktabs}
\usepackage{enumitem}
\setlist[enumerate]{itemsep=0.2ex, parsep=0pt, topsep=0pt, partopsep=0pt}
\setlist[itemize]{itemsep=0.2ex, parsep=0pt, topsep=0pt, partopsep=0pt}

\usepackage{tikz}
\usetikzlibrary{arrows}
\usetikzlibrary{arrows.meta}
\usepackage{tikzscale}

\RestyleAlgo{ruled}
\LinesNumbered
\SetAlgoLined
\IncMargin{1.2em}

\SetCommentSty{mycommfont}
\DontPrintSemicolon
\makeatletter
\newcommand{\clabel}[1]{%
  \addtocounter{AlgoLine}{-1}
  \refstepcounter{AlgoLine}
  \label{#1}
}
\makeatother

\newtheorem{assumption}[theorem]{Assumption}
\makeatletter
\let\c@example\c@theorem
\makeatother

\usepackage{xparse}

\DeclareMathOperator{\anop}{an}
\NewDocumentCommand{\an}{o m}{
  \anop
  \IfValueT{#1}{_{#1}}%
  \!\left(#2\right)%
}

\DeclareMathOperator{\ianop}{\mathcal{I}-an}
\NewDocumentCommand{\ian}{o m}{
  \ianop
  \IfValueT{#1}{_{#1}}%
  \!\left(#2\right)%
}

\DeclareMathOperator{\deop}{de}
\NewDocumentCommand{\de}{o m}{
  \deop
  \IfValueT{#1}{_{#1}}%
  \!\left(#2\right)%
}

\DeclareMathOperator{\ccop}{cc}
\NewDocumentCommand{\cc}{o m}{
  \ccop
  \IfValueT{#1}{_{#1}}%
  \!\left(#2\right)%
}

\DeclareMathOperator{\chop}{ch}
\NewDocumentCommand{\ch}{o m}{
  \chop
  \IfValueT{#1}{_{#1}}%
  \!\left(#2\right)%
}

\DeclareMathOperator{\maop}{ma}
\NewDocumentCommand{\ma}{o m}{
  \maop
  \IfValueT{#1}{_{#1}}%
  \!\left(#2\right)%
}

\DeclareMathOperator{\paop}{pa}
\NewDocumentCommand{\pa}{o m}{
  \paop
  \IfValueT{#1}{_{#1}}%
  \!\left(#2\right)%
}

\newcommand{\indep}{\perp\!\!\!\perp}

\clearauthor{%
 \Name{Francisco Madaleno}\\
 \addr Department of Technology, Management and Economics, Danish Technical University
 \AND
 \Name{Pratik Misra}\\
 \addr Department of Mathematics and Statistics, Binghamton University, State University of New York
 \AND
 \Name{Alex Markham} \Email{\href{mailto:alex.markham@causal.dev}{alex.markham@causal.dev}}\\
 \addr Department of Mathematical Sciences, University of Copenhagen%
}

\hypersetup{colorlinks,citecolor=blue,linkcolor=magenta}

\begin{document}
\maketitle

\begin{abstract}
	Directed acyclic graphical (DAG) models are a powerful tool for representing causal relationships among jointly distributed random variables, especially concerning data from across different experimental settings.
	However, it is not always practical or desirable to estimate a causal model at the granularity of given features in a particular dataset.
	There is a growing body of research on \emph{causal abstraction} to address such problems.
	We contribute to this line of research by
	(i) providing novel graphical identifiability results for practically-relevant interventional settings,
	(ii) proposing an efficient, provably consistent algorithm for directly learning abstract causal graphs from interventional data with unknown intervention targets, and
	(iii) uncovering theoretical insights about the lattice structure of the underlying search space, with connections to the field of causal discovery more generally.
	As proof of concept, we apply our algorithm on synthetic and real datasets with known ground truths, including measurements from a controlled physical system with interacting light intensity and polarization.
\end{abstract}


\section{Introduction}
\label{sec:intro}

Discovering causal relationships is one of the fundamental goals of scientific research.
Nevertheless, the causal relationships of interest are not always between features of a given dataset.
For example, in neuroscience, one may have data describing the interactions of groups of (or even individual) neurons and wish to learn from this data a causal model over cognitive or behavioral states \citep{grosse2024neuro}.
Similarly, in genomics, one may observe gene expression profiles for individual cells, while the goal is to learn a causal model over clusters of cells with similar regulatory states \citep{squires2022causal}.

This is a very general and difficult problem, so we restrict our interest here to particular cases where there is data from some assumed low-level or fine-grained causal DAG model, and we offer a formalization of what it means for a DAG with fewer nodes to be a high-level or coarse-grained abstraction of this underlying causal DAG model.
Our approach is based on the intuition that variables with similar causes and similar effects (graphically, similar ancestors and descendants) are in some sense redundant and can be abstracted away by clustering them together while retaining only the salient causal relationships.

There is a growing number of similarly motivated works on consistent transformations of causal models \citep{rubensteincausal} and causal abstractions \citep{BeckersH19,beckers2020approximate,beckers2021equivalent,otsuka2022equivalence} that lay a groundwork for formally describing such problems.
There is also some work on learning abstractions \citep{massidda2024learning}, on directly learning a reduced DAG according to its marginal independences \citep{astat}, on using partitions to simplify learning large graphs \citep{gu2020learning}, and on clustering nodes in a DAG while preserving the identifiability of specified causal estimands \citep{tikka2023clustering}.
Other approaches focus on causal layerings \citep{feigenbaum2024causal}, on learning abstractions in the unsupervised setting \citep{zhu2024unsupervised}, and on targeted causal reduction for extracting compact high-level explanations of simulator phenomena and of reinforcement learning policy behavior \citep{kekic2024targeted,kekic2025nonlinear}.
Similarly, for complex decision settings, \citet{dyer2024ics} develop interventionally consistent surrogates for expensive simulators, while \citet{dyer2025bandit} accelerate decision making in causal bandit problems by enabling transfer across levels of granularity.
A broader discussion of related literature is provided in \Cref{app:related-work}.

In contrast to existing work, our main contribution is to develop a general framework for understanding causal abstractions graphically, which we call coarsening.
In particular, we study the space of all coarsenings, proving that it has ``nice'' mathematical structure.
This allows us:
(i) to provide novel graphical identifiability results, in particular for practically-relevant interventional coarsenings, also suggesting new directions for future work on abstractions;
(ii) to propose an efficient, provably consistent algorithm for learning interventional coarsenings directly from data---a coarsening is a simpler mathematical object than a fine-grained DAG model, and our algorithm leverages this;
and (iii) to establish the fundamental mathematical properties of the coarsening space, connecting it to the search space of other causal discovery algorithms.

The paper proceeds as follows:
\Cref{sec:types-coarsening} formalizes the general problem of graphical causal abstraction, establishes the lattice structure of the coarsening space, establishes identifiability of interventional coarsenings, and draws connections to causal discovery more broadly.
\Cref{sec:learn-coars-caus} develops a flexible learning algorithm for interventional coarsenings, with formal consistency and complexity guarantees.
All proofs are deferred to \Cref{app:proofs-results-main}.
For empirical evaluation, \Cref{sec:experimental-results} presents an open-source implementation of the algorithm, evaluating it on synthetic data to empirically test the theoretical guarantees and on real data, compared to baselines from the literature, to demonstrate practical relevance.
\Cref{sec:discussion} concludes with a discussion of our approach, its limitations, and future directions.

\section{Coarsening Causal DAGs}
\label{sec:types-coarsening}

The following section relies on basic ideas from graphical models~\citep{steffen}, order theory~\citep{davey2002introduction}, and combinatorics~\citep{stanley2011enumerative}.

For a DAG \(G=(V,E)\), in addition to the familiar terminology of parents \(\pa[G]{v}\) and children \(\ch[G]{v}\), we denote \emph{ancestors} as \(\an[G]{v} \coloneqq \{w \in V \mid\) there exists a directed path from \(w \text{ to } v\}\) and analogously-defined \emph{descendants} as \(\de[G]{v}\)---note that every node is its own ancestor and descendant.
These all extend naturally to sets: the parents of a set of nodes is the union of the parent sets of the nodes, and so on.
A distribution is \emph{Markov} to a DAG if the distribution's conditional independence statements are implied by the DAG's \(d\)-separation statements---the set of all such distributions for a given DAG is denoted \(\mathcal{M}(G)\).

A \emph{poset} (partially ordered set) is a set with a reflexive, transitive, and antisymmetric order relation \(\preceq\); a \emph{lattice} is a poset where every pair of elements has a unique meet (greatest lower bound) and join (least upper bound).

A \emph{partition} of \(V\) is a collection \(\Pi = \{\pi_1, \ldots, \pi_k\}\) of disjoint non-empty sets, called \emph{parts}, whose union is \(V\).
A partition \(\Pi\) \emph{refines} another \(\Pi'\) (written \(\Pi\preceq\Pi'\)) if every part of \(\Pi\) is contained in some part of \(\Pi'\)---equivalently, \(\Pi'\) is said to \emph{coarsen} \(\Pi\).
The collection of all partitions of \(V\), ordered by refinement, forms the \emph{partition refinement lattice}, where the meet of two partitions is their finest common refinement and the join is their coarsest common coarsening.

\subsection{General Coarsenings}
\label{sec:general-coarsenings}

We begin with what we argue are necessary but not sufficient graphical conditions for a ``reasonable'' causal abstraction:

\begin{definition}
	\label[definition]{def:gen-coarse}
	Given a DAG \(G = (V, E)\), a \emph{coarsening} is a DAG \(G' = (V', E')\) for which there exists a surjection \(\chi: V \rightarrow V'\) such that
	\[E' = \{\chi(v) \to \chi(w) \mid v\to w \in E, \chi(v) \neq \chi(w)\}.\]
\end{definition}
For example, this precludes coarsenings that reverse causal edges as well as coarsenings that group \(\{u,w\}\) together apart from \(\{v\}\) in graph structures like \(u \rightarrow v \rightarrow w\).
More generally, it ensures model containment:

\begin{lemma}
	\label[lemma]{lem:imap}
	Let $G$ be a DAG and $G'$ be one of its coarsenings.
	Every distribution Markov to \(G\) is also Markov to \(G'\), that is, \(\mathcal{M}(G) \subseteq \mathcal{M}(G')\).
\end{lemma}
This containment induces a natural partial order over coarsenings, forming a lattice:

\begin{theorem}
	\label{thm:lattice}
	For any DAG \(G=([d],E)\), the poset of its coarsenings is a lattice, more specifically, a sublattice of the partition refinement lattice of $[d]$.
\end{theorem}

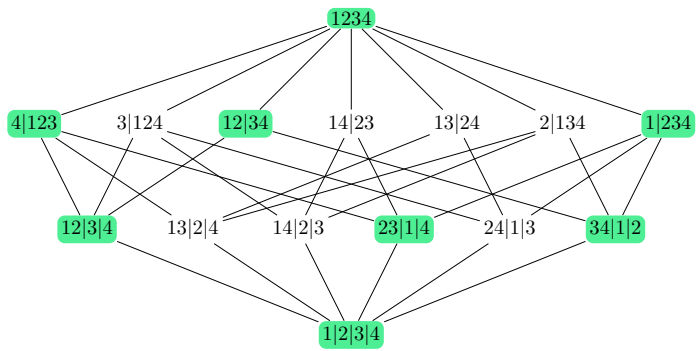
\begin{wrapfigure}{r}{0.61\textwidth}
	\centering
	\vspace{-10pt}
	\begin{tikzpicture}[
		every node/.style={fill=white, inner sep=2pt},
	]

	\node[fill=SeaGreen2,rounded corners] (1234) at (0,0) {1234};

	\node[fill=SeaGreen2,rounded corners] (4_123) at (-6,-2) {123\textbar 4};
	\node (3_124)                      at (-4,-2) {124\textbar 3};
	\node[fill=SeaGreen2,rounded corners] (12_34) at (-2,-2) {12\textbar 34};
	\node (14_23)                      at ( 0,-2) {14\textbar 23};
	\node (13_24)                      at ( 2,-2) {13\textbar 24};
	\node (2_134)                      at ( 4,-2) {134\textbar 2};
	\node[fill=SeaGreen2,rounded corners] (1_234) at ( 6,-2) {1\textbar 234};

	\node[fill=SeaGreen2,rounded corners] (12_3_4) at (-5,-4) {12\textbar 3\textbar 4};
	\node (13_2_4)                      at (-3,-4) {13\textbar 2\textbar 4};
	\node (14_2_3)                      at (-1,-4) {14\textbar 2\textbar 3};
	\node[fill=SeaGreen2,rounded corners] (23_1_4) at ( 1,-4) {1\textbar 23\textbar 4};
	\node (24_1_3)                      at ( 3,-4) {1\textbar 24\textbar 3};
	\node[fill=SeaGreen2,rounded corners] (34_1_2) at ( 5,-4) {1\textbar 2\textbar 34};

	\node[fill=SeaGreen2,rounded corners] (singletons) at (0,-6)
	{1\textbar 2\textbar 3\textbar 4};

	\foreach \i in {12_34,13_24,14_23,1_234,2_134,3_124,4_123}
	\draw (1234) -- (\i);

	\draw (12_34) -- (12_3_4);
	\draw (12_34) -- (34_1_2);

	\draw (13_24) -- (13_2_4);
	\draw (13_24) -- (24_1_3);

	\draw (14_23) -- (14_2_3);
	\draw (14_23) -- (23_1_4);

	\draw (1_234) -- (23_1_4);
	\draw (1_234) -- (24_1_3);
	\draw (1_234) -- (34_1_2);

	\draw (2_134) -- (13_2_4);
	\draw (2_134) -- (14_2_3);
	\draw (2_134) -- (34_1_2);

	\draw (3_124) -- (12_3_4);
	\draw (3_124) -- (14_2_3);
	\draw (3_124) -- (24_1_3);

	\draw (4_123) -- (12_3_4);
	\draw (4_123) -- (13_2_4);
	\draw (4_123) -- (23_1_4);

	\foreach \i in {12_3_4,13_2_4,23_1_4,24_1_3,14_2_3,34_1_2}
	\draw (\i) -- (singletons);

\end{tikzpicture}
	\captionof{figure}{The partition refinement lattice on 4 nodes, with an example coarsening sublattice \textcolor{SeaGreen3}{highlighted}.}
	\label{fig:example-lattice}
	\vspace{-20pt}
\end{wrapfigure}

To make this more concrete, consider the following example:

\begin{example}
	\label[example]{ex:lattice-simple}
	Let $G=(V,E)$ be the DAG $1\rightarrow 2 \rightarrow 3 \rightarrow 4$ and consider the partition lattice of $[4]$ shown in \Cref{fig:example-lattice}.
	The lattice is organized by the number of parts: level $i$ (for $i \in \{0,1,2,3\}$) contains all partitions of $[d]$ (here, $d=4$) into $i+1$ parts, counted by the \emph{Stirling numbers of the second kind} $S(d, i+1)$.
	These satisfy the recurrence $S(n,k) = k \cdot S(n-1,k) + S(n-1,k-1)$.
	For $d=4$, this gives: $S(4,1)=1$, $S(4,2)=7$, $S(4,3)=6$, and $S(4,4)=1$.
	The total number of partitions is the \emph{Bell number} $B_n = \sum_{k=1}^{n} S(n,k)$.

	Not every partition corresponds to a valid coarsening of $G$.
	Consider the partition $1|2|34$, represented by coarse nodes $V'=\{1,2,\{3,4\}\}$\footnote{To keep notation light, we conflate singleton sets and their elements, e.g., writing \(1\) instead of \(\{1\}\), when it is clear from context.}.
	This is a valid coarsening: the surjection $\chi$ defined by $\chi(1)=1, \chi(2)=2, \chi(3)=\chi(4)=\{3,4\}$ induces the coarsened DAG $G'$ with $1\rightarrow 2 \rightarrow \{3,4\}$.
	In contrast, the partition $1|3|24$ (corresponding to $V''=\{1,3,\{2,4\}\}$) does \emph{not} correspond to a valid coarsening.
	Although a surjection to $V''$ exists, the edges it induces do not form a DAG: \(2 \xrightarrow{G} 3\) requires \(\chi(2) \xrightarrow{G''}\chi(3)\) while \(3 \xrightarrow{G} 4\) requires \(\chi(3) \xrightarrow{G''}\chi(4)=\chi(2)\), forming a cycle.
\end{example}

\begin{algorithm2e}[t]

	\SetKwFunction{RePaRe}{RePaRe}
	\SetKwFunction{Refine}{Refine}
	\SetKwFunction{IsEdge}{IsEdge}
	\SetKwProg{Fn}{Function}{:}{}

	\KwIn{Coarsening $G = (\Pi, E)$}
	\KwOut{Refinement $G^*$}

	\Fn{\RePaRe{$G$}}{
		\tcp{Attempt split}
		$(\pi^*, \{\pi_a, \pi_b\}) \coloneqq$ \Refine{$\Pi$} \clabel{ln:refine} \;
		\lIf{$\pi^* = \emptyset$}{\Return $G$}

		\tcp{Define new node set}
		$\Pi' \coloneqq (\Pi \setminus \{\pi^*\}) \cup \{\pi_a, \pi_b\}$ \clabel{ln:new-part}\;

		\tcp{Define new edge set}
		$
			\begin{aligned}
				E' \coloneqq \; & \{ u \to v \in E \mid u \neq \pi^* \text{ and } v \neq \pi^* \}                                                                                                \\
				                & \cup \;         \{ u \to v \mid \{u, v\} = \{\pi_a, \pi_b\}, \mathtt{IsEdge}(u, v)  \}                                                                         \\
				                & \cup \;         \{ \pi \to \pi_{\mathrm{new}} \mid \pi \in \pa[G]{\pi^*}, \pi_{\mathrm{new}} \in \{\pi_a, \pi_b\}, \mathtt{IsEdge}(\pi, \pi_{\mathrm{new}}) \} \\
				                & \cup \;         \{ \pi_{\mathrm{new}} \to \pi \mid \pi \in \ch[G]{\pi^*}, \pi_{\mathrm{new}} \in \{\pi_a, \pi_b\}, \mathtt{IsEdge}(\pi_{\mathrm{new}}, \pi) \}
			\end{aligned}
		$ \clabel{ln:is-edge} \;

		\tcp{Construct and Recurse}
		$G' \coloneqq (\Pi', E')$\;
		\Return \RePaRe{$G'$}\;
	}
	\caption{Recursive Partition Refinement for DAG Learning}
	\label{alg:repare}
\end{algorithm2e}

The lattice structure means that the space of coarsenings is a ``nice'' space to work with in the sense that we can traverse it easily with something like \Cref{alg:repare}.
This gives us a framework for identifying \emph{any} valid coarsening by choosing suitable oracles for \Refine (\Cref{ln:refine}) and \IsEdge (\Cref{ln:is-edge}) functions.

\begin{definition}
	\label[definition]{def:oracles}
	Given an underlying DAG \(\underline{G} = ([d],\underline{E})\) and a coarsening \(G = (\Pi,E)\), a \emph{\Refine-oracle} is any function that takes in \(\Pi\) and returns a part \(\pi^*\in\Pi\) that itself is partitioned into \(\pi_a, \pi_b\) satisfying:
	\begin{enumerate}
		\item \emph{acyclicity preservation}: there does not exist a directed cycle in \(G\) between the sets \(\pi_a\) and \(\pi_b\), i.e., either \(\an[\underline{G}]{\pi_a} \cap \pi_b = \emptyset\) or \(\an[\underline{G}]{\pi_b} \cap \pi_a = \emptyset\).
	\end{enumerate}
	If no such \(\pi^*\) exists, the oracle returns \((\emptyset, \emptyset)\).
	Furthermore, an \emph{\IsEdge-oracle} is any function taking in parts \(u,v\in\Pi'\) (\Cref{ln:new-part}) and returning \texttt{True} if and only if \(u,v\) satisfy:
	\begin{enumerate}[resume]
		\item \emph{parent consistency}: \(\pa[\underline{G}]{v} \cap u \neq \emptyset\).
	\end{enumerate}
\end{definition}

\begin{theorem}[Completeness of \RePaRe]
	\label{thm:completeness}
	Let \(\underline{G} = (\underline{V}, \underline{E})\) be a DAG and \(G^*\) be any valid coarsening of \(\underline{G}\).
	There exist \Refine- and \IsEdge-oracles (\Cref{def:oracles}) such that \Cref{alg:repare}, with the trivial coarsening \(\overline{G}=(\{V\}, \emptyset)\) as input, terminates and outputs \(G^*\).
\end{theorem}

The proof constructs explicit oracles by leveraging the lattice structure of valid coarsenings.
Whenever $\Pi^* \prec \Pi$, we can find a part $\pi \in \Pi$ that must be split in any refinement chain from $\Pi$ to $\Pi^*$.
The \Refine-oracle identifies such a part and splits it according to the structure imposed by $\Pi^*$, while the \IsEdge-oracle determines edges in the refined coarsening by querying the true parental relations of $\underline{G}$.
This greedy refinement is guaranteed to converge to $\Pi^*$ because every step moves strictly closer in the lattice order.

In the next subsections, we show how this general notion of coarsening can be narrowed down into other (perhaps more causally- or domain-relevant) notions of coarsening, but first we look at an example of how the lattice is already interesting for understanding existing---and perhaps developing future---causal discovery algorithms:

\begin{example}
	\label[example]{ex:gen-coarse-lingam}
	Let $G$ be the 4-node path $1 \rightarrow 2 \rightarrow 3 \rightarrow 4$.
	A topological order $\sigma=(1,2,3,4)$ induces a natural refinement chain in the
	coarsening lattice: \(1234 \succ 1|234 \succ 1|2|34 \succ 1|2|3|4\).
	This illustrates how causal discovery methods that estimate a topological order can be interpreted as walking down the lattice.
	For example, under linear non-Gaussian assumptions, DirectLiNGAM~\citep{directLiNGAM} identifies a causal ordering from observational data and thereby a compatible DAG---its search corresponds to such a walk.
\end{example}

\subsection{Interventional Coarsening}
\label{sec:interv-coars}

We now narrow down the general notion of coarsening from \Cref{def:gen-coarse}, giving it a more rigorous causal grounding in interventions and demonstrating how the general framework can be adapted to answer meaningful practical questions.
Across scientific domains, a natural question is: \emph{given a collection of post-intervention samples, which of the measured features are affected similarly, and what is the salient causal structure among these clusters?}\footnote{This question mirrors common practice in high-dimensional interventional settings, where one summarizes responses by clusters and studies relations between them.
	In genomics, \textsc{CWGCNA} inserts a mediation-based causal inference step into WGCNA to estimate causal relationships among phenotypes, gene modules, and module features \citep{liu2024cwgcna}.
	In neuroscience, TMS-EEG analyses track how stimulation-evoked activity propagates through brain networks \citep{xiao2025tms}.
	In simulation modeling, Targeted Causal Reduction learns a concise set of causal factors explaining a specified target phenomenon from interventional simulation data \citep{kekic2024targeted}, and in robotics \textsc{SCALE} uses simulated interventions to discover causally relevant context variables associated with distinct manipulation modes \citep{lee2023scale}.}

Thus, we begin by formalizing the idea of ``measured features being affected similarly''.
Relative to a set of (possibly soft) interventions \(\mathcal{I}\), denote a node's \emph{intervened ancestors} as \(\ian[G]{v} \coloneqq \{ w \in \an[G]{v} \mid \exists I \in \mathcal{I} \text{ such that } w \in I\}.\)
This naturally leads to interventional coarsening:

\begin{definition}
	\label[definition]{def:ivn-coarse}
	Given a DAG \(G=(V,E)\) and a set of interventions \(\mathcal{I}\), define the \emph{interventional coarsening}, denoted \(G^{\mathcal{I}}\), to be the coarsening whose partition surjection \(\chi\) satisfies
	\[\text{for all } v,w \in V,\quad\chi(v) = \chi(w) \iff \ian[G]{v} = \ian[G]{w}.\]
\end{definition}

Framed in this way, it is clear that \(G^{\mathcal{I}}\) is a valid coarsening and thus one that is identifiable according to the completeness result of \Cref{thm:completeness} by specifying appropriate oracles.
In this case, the oracles need to be able to work with distributions rather than merely graphs as well as to be able to determine which nodes are descendants of which interventions---formally:

\begin{assumption}
	Given a ground-truth DAG \(G = (V, E)\) and intervention set \(\mathcal{I}\) (which includes \(\emptyset\), representing the observational setting), we make the following assumptions about the associated distributions:
	\label[assumption]{asm:ivn-coarse}

	\begin{enumerate}
		\item \emph{coarse Markov:} the distributions factorize according to \(G^{\mathcal{I}}\), i.e., for all \(I \in \mathcal{I}\):
		      \[f^I(X_V) = \prod_{u\in \chi(V)\setminus\chi(I)} f^{\emptyset}\big(X_{\chi^{-1}(u)}\mid X_{\chi^{-1}(\pa[G^{\mathcal{I}}]{u})}\big) \prod_{u\in \chi(I)} f^I\big(X_{\chi^{-1}(u)}\mid X_{\chi^{-1}(\pa[G^{\mathcal{I}}]{u})}\big).\]

		\item \emph{coarse faithfulness:} the only conditional independences among coarse nodes \(\chi(V)\) are those implied by the Markov factorization above.
		\item \emph{interventional soundness:} interventions induce changes in marginal distributions compared to the observational distribution, that is:
		      \[\text{for all } I \in \mathcal{I}\setminus\emptyset \text{ and } v \in V,\quad v \in \de[G]{I} \implies f^I(X_v) \neq f^{\emptyset}(X_v).\]
	\end{enumerate}
\end{assumption}

Note that, when \(\chi(V) \neq V\), these are strictly weaker than the usual Markov and faithfulness assumptions---they allow violations at the \(V\)-level as long as those do not induce violations at the \(\chi(V)\)-level.
Making use of these assumptions, we can specify oracles that plug into \Cref{alg:repare} to enable us to prove identifiability of \(G^{\mathcal{I}}\):

\begin{theorem}
	\label{thm:ivn-coarse}
	Under \Cref{asm:ivn-coarse}, the interventional coarsening \(G^{\mathcal{I}}\) is identifiable.
\end{theorem}

As we will see in \Cref{sec:learn-coars-caus}, replacing the oracles used to prove \Cref{thm:ivn-coarse} with valid statistical hypothesis tests allows us to turn \Cref{alg:repare} into a consistent estimator of \(G^{\mathcal{I}}\).
However, we first spend the rest of this section building up theoretical intuitions with the coarsening framework.

\begin{example}
	\label[example]{ex:ivn-coarse}
	Let $G$ be the DAG on $V=\{1,2,3,4\}$ with edges $1\to 3$, $2\to 3$, and $3\to 4$.
	Consider the intervention set $\mathcal{I}=\{\emptyset,\{1\},\{2\}\}$, where $\emptyset$ denotes the observational regime and $\{1\},\{2\}$ are single-target interventions.

	For each node $v$, $\ian[G]{v}$ collects the intervened ancestors of $v$.
	This yields
	\[
		\ian[G]{1}=\{1\},\quad
		\ian[G]{2}=\{2\},\quad
		\ian[G]{3}=\{1,2\},\quad
		\ian[G]{4}=\{1,2\}
	\]
	Thus the interventional coarsening groups nodes that are affected by the same intervention targets, yielding the partition \(\Pi^{\mathcal{I}}=\bigl\{\{1\},\{2\},\{3,4\}\bigr\}\).
	The induced coarsened DAG has two edges $\{1\}\to\{3,4\}$ and $\{2\}\to\{3,4\}$.

	This example highlights how $\mathcal{I}$ determines the \emph{resolution} of the learnable coarse structure: descendants that are indistinguishable with respect to the available interventions are merged.
	This intervention-driven notion of abstraction is closely aligned with the role	of distributional changes across environments in causal discovery with unknown targets, as in UT-IGSP~\citep{squires2020permutation} and GnIES~\citep{gamella2022characterization}.
\end{example}

\subsection{Other Coarsenings}
\label{sec:other-coarsenings}

In contrast to the general and interventional coarsenings above, we now present two specific observational coarsenings, one related to Markov equivalence classes~\citep{AMP97} and one related to unconditional equivalence classes~\citep{astat}.

\subsubsection{Essential Coarsenings}
\label{sec:essent-coars}

Classical identifiability for causal discovery in the observational setting extends only to Markov equivalence classes (MECs) of DAGs.
\citet{AMP97} represent these equivalence classes with \emph{essential graphs} (CPDAGs), whose nodes can be partitioned into chain components (which we denote \(\cc[G]{v}\)), and whose edges can be partitioned into directed and undirected sets.
The directed edges represent all the direct causal relations that are identifiable without interventions or further assumptions, and the undirected edges encode uncertainty about the causal direction; an edge is directed if and only if it is between two different chain components.
Our general notion of coarsening (\Cref{def:gen-coarse}) can be narrowed down to capture much of the same information:

\begin{definition}
	\label[definition]{def:ess-coarse}
	Given a DAG \(G\), its \emph{essential coarsening} is the coarsening with partition surjection \(\chi\) defined by:
	\[\text{for all } v,w \in V^G,\quad\chi(v) = \chi(w) \iff \cc[G]{v} = \cc[G]{w}.\]
\end{definition}

\begin{example}
	\label[example]{ex:ess-coarse}
	Let $G$ be the DAG on $V=\{1,2,3,4,5\}$ with edges $1\to 2$, $2\to 3$, $4\to 3$, and $3\to 5$.
	The MEC of $G$ is represented by a CPDAG whose edges are $1-2$ (undirected), $2\to 3 \leftarrow 4$ (directed) and $3\to 5$ (directed), since the directed edges incident to $3$ are strongly protected while the one between $1$ and $2$ is not \citep{AMP97}.

	The chain components of the CPDAG are \(\{1,2\}, \{3\}, \{4\}, \{5\}\), and the essential coarsening is given by the partition \(\Pi^{\mathrm{ess}} = \bigl\{\{1,2\},\{3\},\{4\},\{5\}\bigr\}\) with induced coarsened edges $\{1,2\}\to \{3\}$, \(\{3\}\rightarrow\{5\}\), and $\{4\}\to \{3\}$.
	Hence, the essential coarsening preserves the chain components and the partial order between but not the undirected structure within each chain component.

	From this perspective, a purely observational algorithm analogous to PC~\citep{kalisch2007estimating} or GES~\citep{chickering2002optimal} could use conditional independences to move down the coarsening lattice until arriving at the essential coarsening, at which point no further refinement is justified by observational data.
	Including background knowledge~\citep{bang2023we} could substantially reduce the search space within the lattice.
\end{example}

\subsubsection{Marginal Coarsenings}
\label{sec:marginal-coarsenings}

Unconditional equivalence classes (UECs) provide a coarser notion of equivalence than MECs, capturing less causal information---still identifying all v-structures and possible source nodes---but at a much cheaper cost:
a UEC can be learned via quadratically-many pairwise \emph{marginal} independence tests rather than exponentially-many conditional independence tests.
UECs are also relevant for latent causal DAGs \citep{pmlr-v124-markham20a,markham2023neurocausalfactoranalysis,jiang2023learning} as well as causal clustering \citep{pmlr-v177-markham22a}.
\citet[Theorem~1~(2)]{pmlr-v186-markham22a} provide a characterization of UECs that relies on grouping nodes together when they share the same set of maximal ancestors (denoted \(\ma[G]{v}\)).
Using this, our general notion of coarsening (\Cref{def:gen-coarse}) can be narrowed down to encode exactly the information captured by a UEC:

\begin{definition}
	\label[definition]{def:marg-coarse}
	Given a DAG $G$, its \emph{marginal coarsening} is the coarsening with partition surjection $\chi$ defined by:
	\[\text{for all } v,w \in V^G,\quad\chi(v) = \chi(w) \iff \ma[G]{v} = \ma[G]{w}.\]
\end{definition}

\begin{example}
	\label[example]{ex:marginal}
	For the DAG $G$ from \Cref{ex:ess-coarse}, its UEC is represented by the DAG-reduction $\{1,2\}\to \{3,5\} \leftarrow \{4\}$~\citep[Definition~5.1.2]{astat}, corresponding to the partition $12|4|35$.
	This is indeed a marginal coarsening: nodes $3$ and $5$ have the same maximal ancestors $\ma[G]{3}=\ma[G]{5}=\{1,4\}$, while $\ma[G]{4}=\{4\}$ and \(\ma[G]{1}=\ma[G]{2}=\{1\}\).
	Thus, the reduced DAG can be obtained by traversing the coarsening lattice downward via marginal independence tests.
\end{example}

\section{Learning Interventional Coarsenings}
\label{sec:learn-coars-caus}

In place of the oracles used for identifiability in \Cref{thm:ivn-coarse}, we now present practical algorithms that turn \Cref{alg:repare} from a theoretical tool into a method for learning interventional coarsenings from data.
For simplicity, we assume Gaussianity and employ two standard statistical tests.\footnote{Note that the Gaussianity assumption can easily be relaxed and the test statistics replaced with something more general, like energy distance~\citep{szekely2013energy} or maximum mean discrepancy~\citep{gretton2012kernel} for the two-sample test and distance correlation~\citep{szekely2007measuring} or the Hilbert-Schmidt independence criterion~\citep{gretton2007kernel} for the independence test.}

\begin{algorithm2e}[t]
	\SetKwFunction{RefineAux}{RefineAux}
	\SetKwProg{Fn}{Function}{:}{}

	\KwIn{Data $\mathcal{D} = \{ X^{(I)} \}_{I \in \mathcal{I}}$, Threshold $\alpha$}
	\KwOut{Intervention descendant indicator matrix $M \in \{0,1\}^{|V| \times |\mathcal{I}|}$}

	\Fn{\RefineAux{$\mathcal{D}, \alpha$}}{
	\For{$(v, I) \in V \times \mathcal{I}$ \textbf{in parallel}}{
	$p_{v,I} \coloneqq \text{Welch}_t\!\left( X_v^{(I)}, X_v^{(\emptyset)} \right)$ \;
	$M_{v,I} \coloneqq \mathbb{1}[p_{v,I} < \alpha]$ \;
	}
	\Return $M$\;
	}
	\caption{Precompute intervention descendants}
	\label{alg:refine-aux}
\end{algorithm2e}

\begin{algorithm2e}[t]
	\SetKwFunction{RefineTest}{RefineTest}
	\SetKwProg{Fn}{Function}{:}{}
	\KwIn{Partition $\Pi$, Descendant patterns $M$}
	\KwOut{Pair $(\pi^*, \{\pi_a, \pi_b\})$ or $(\emptyset, \emptyset)$}

	\Fn{\RefineTest{$\Pi, M$}}{
		\For{\textbf{each} $\pi \in \Pi$ \textbf{with} $|\pi| > 1$}{
			Pick $u \in \pi$\;
			$\pi_a \coloneqq \{ v \in \pi \mid M_{v,\cdot} = M_{u,\cdot} \}$\;
			$\pi_b \coloneqq \pi \setminus \pi_a$\;
			\lIf{$\pi_b \neq \emptyset$}{\Return $(\pi, \{\pi_a, \pi_b\})$}
		}
		\Return $(\emptyset, \emptyset)$\;
	}
	\caption{Refine via intervention descendant patterns}
	\label{alg:refine-test}
\end{algorithm2e}

\begin{algorithm2e}[t]
	\SetKwFunction{IsEdgeTest}{IsEdgeTest}
	\SetKwProg{Fn}{Function}{:}{}
	\KwIn{\small Parts $U, W \subseteq V$, Observational data $X$, Conditioning set $Z \subseteq V$, Threshold $\alpha$}
	\KwOut{\texttt{True} or \texttt{False}}

	\Fn{\IsEdgeTest{$U, W, X, Z, \alpha$}}{
		$\tilde{X}_U \coloneqq X_U - \mathbb{E}[X_U \mid X_Z]$, $\tilde{X}_W \coloneqq X_W - \mathbb{E}[X_W \mid X_Z]$ \tcp*{Regression residuals}
		$\Lambda \coloneqq \mathrm{Wilks}_{\lambda}\!\left( \tilde{X}_U, \tilde{X}_W \right)$ \tcp*{CCA-based test statistic}
		$p \coloneqq p\text{-value}(\Lambda)$ \tcp*{Under $H_0: X_U \indep X_W \mid X_Z$}
		\Return $p < \alpha$\;
	}
	\caption{IsEdge via conditional CCA and Wilks' Lambda}
	\label{alg:is-edge-test}
\end{algorithm2e}

First, \RefineAux (\Cref{alg:refine-aux}) precomputes an \emph{intervention descendant indicator matrix} $M$ using a $t$-test \citep{welch1947generalization} to recover which nodes respond to each intervention.
Then, \RefineTest (\Cref{alg:refine-test}) greedily selects the first part containing nodes with different intervention descendant patterns and splits it.
Finally, \IsEdgeTest\footnote{See the proof of \Cref{thm:ivn-coarse} for details about constructing the conditioning sets \(Z\).} (\Cref{alg:is-edge-test}) applies canonical correlation analysis (CCA) and computes a likelihood ratio test \citep[Chapter~11]{hotelling1936relations,cca} to determine directed edges between coarsened nodes.
Plugging in statistical tests in this way, the identifiability in \Cref{thm:ivn-coarse} leads straightforwardly to consistency (\Cref{cor:asympt-cons}) and allows us to analyze the time complexity of learning interventional coarsenings (\Cref{thm:ivn-complex}).

\begin{corollary}
	\label[corollary]{cor:asympt-cons}
	Given Gaussian distributed \(X_V\) satisfying \Cref{asm:ivn-coarse}, \RePaRe using \RefineTest and \IsEdgeTest learns the correct coarsening in the large sample limit.
\end{corollary}

\begin{theorem}
	\label{thm:ivn-complex}
	Let $d = |V|$ be the number of nodes in the underlying DAG, $e = |\mathcal{I}|$ the number of interventions, and $n = \max_{I \in \mathcal{I}} |X^{(I)}|$ the maximum sample size across datasets.
	Let $k = |\Pi^{\mathcal{I}}|$ be the number of parts in the true coarsening and $p = \max_{\pi \in \Pi^{\mathcal{I}}} |\pi|$ be the maximum size of any part.
	\RePaRe, using \RefineTest and \IsEdgeTest, and assuming the statistical regime where \(d,p\leq n\), runs in worst-case time \(O\!\left(den + k^2 p^2n\right)\).
\end{theorem}

\RePaRe decouples the dependence on underlying graph size $d$ from dependence on coarsened graph complexity.
The precomputation and refinement phases incur $O(den)$ cost (inherent to processing $d$ nodes), but crucially, the expensive edge-finding phase depends only on the coarsened graph size $k$ and part sizes $p$, costing $O(k^2 p^2 n)$.
The algorithm pays for large $d$ once (in refinement), then scales with the simpler coarsened structure.
This makes \RePaRe particularly suited for learning causal structure at ``human scale''---for instance, coarsened graphs with only tens of nodes---remaining efficient regardless of the size of $d$.


\section{Experimental Results}
\label{sec:experimental-results}

Across synthetic and real data, we evaluate whether \RePaRe recovers the ground-truth interventional coarsening, measuring partition recovery with the adjusted Rand index (ARI)~\citep[Eq~(5)]{hubert1985comparing} and edge recovery with the F-score~\citep[Eq~(8.6)]{manning2009introduction}.
In all experiments, we run \RePaRe with the \Cref{sec:learn-coars-caus} algorithms:
\RefineTest for partition refinement, and \IsEdgeTest for edge recovery.\footnote{An open source implementation as well as scripts for reproducing all of the following experiments can be found at \url{https://github.com/Alex-Markham/repare/tree/v0.2.0}.}

\subsection{Synthetic Data}
\label{sec:synthetic-data}

\begin{figure}[t]
	\centering
	\subfigure[partition recovery, \(\iota=2\)]{
		\includegraphics[width=0.3\textwidth]{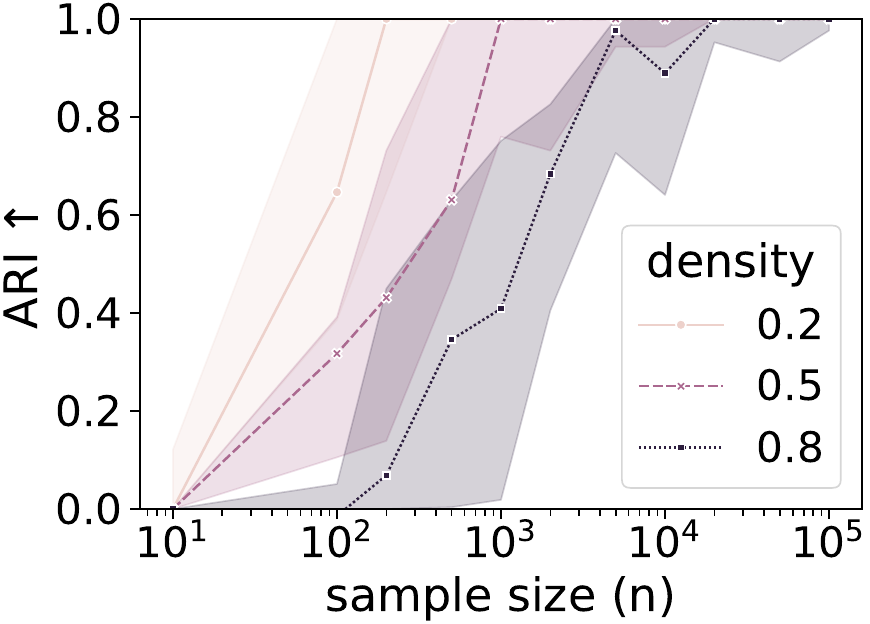}
		\label{fig:ari-k2}
	}
	\hfill
	\subfigure[partition recovery, \(\iota=5\)]{
		\includegraphics[width=0.3\textwidth]{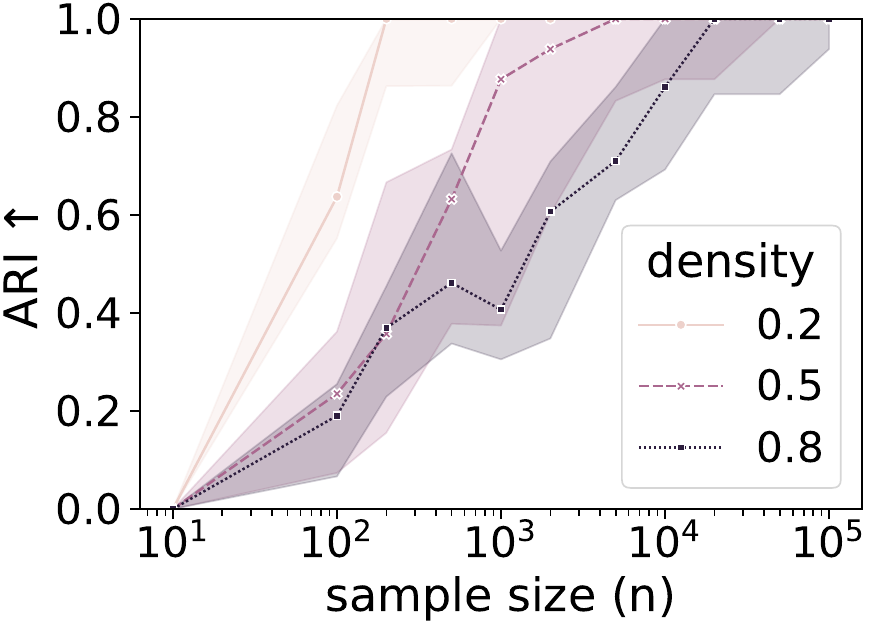}
		\label{fig:ari-k5}
	}
	\hfill
	\subfigure[partition recovery, \(\iota=8\)]{
		\includegraphics[width=0.3\textwidth]{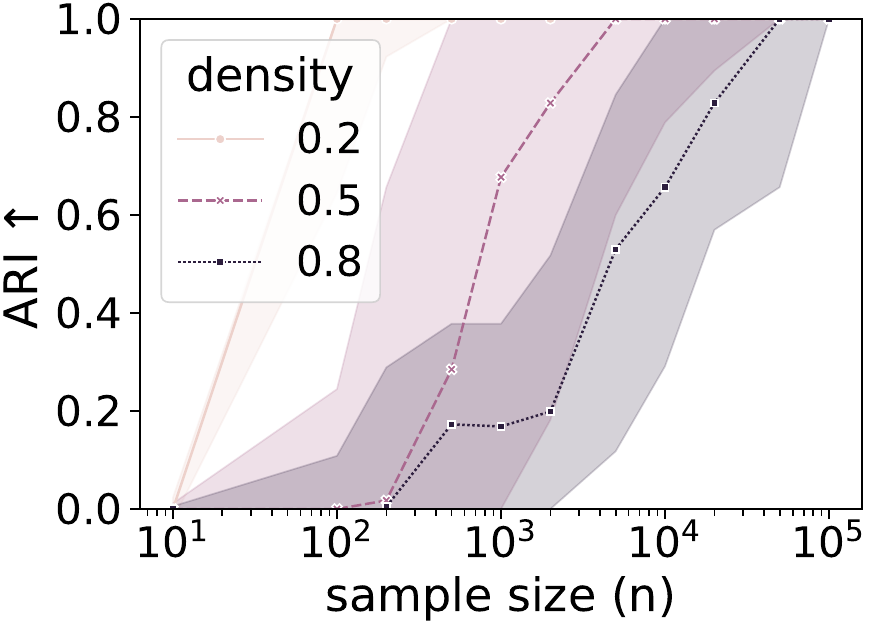}
		\label{fig:ari-k8}
	}

	\medskip

	\subfigure[edge recovery, \(\iota=2\)]{
		\includegraphics[width=0.3\textwidth]{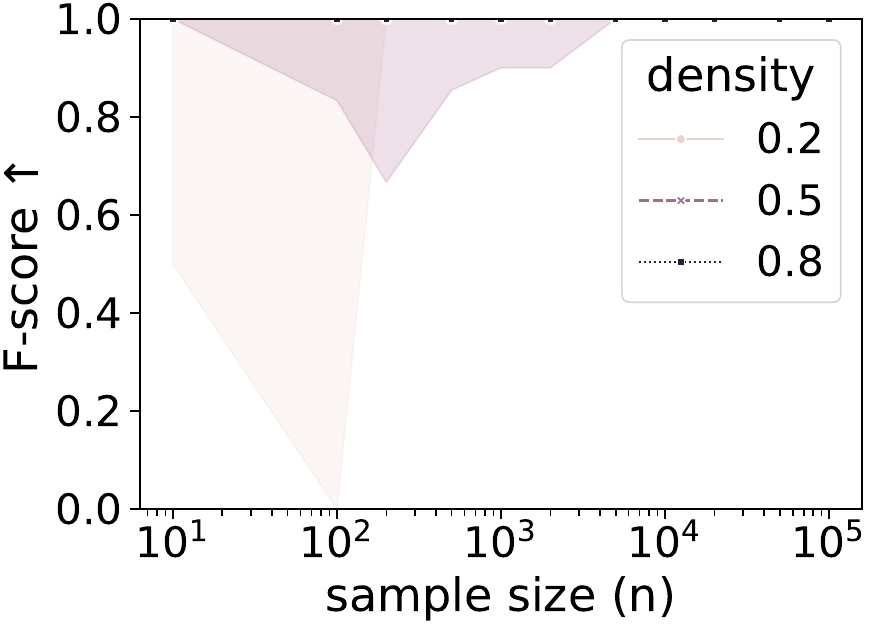}
		\label{fig:fscore-k2}
	}
	\hfill
	\subfigure[edge recovery, \(\iota=5\)]{
		\includegraphics[width=0.3\textwidth]{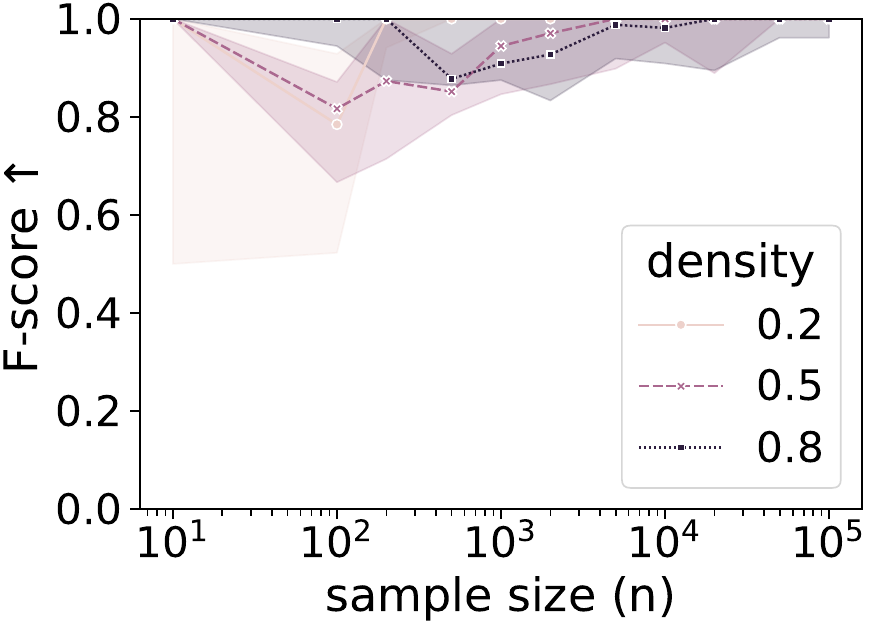}
		\label{fig:fscore-k5}
	}
	\hfill
	\subfigure[edge recovery, \(\iota=8\)]{
		\includegraphics[width=0.3\textwidth]{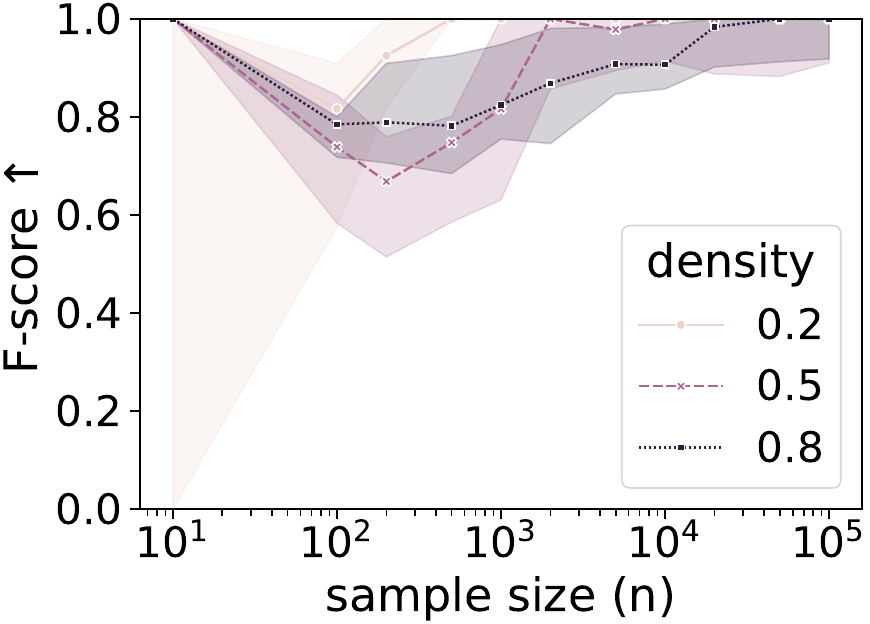}
		\label{fig:fscore-k8}
	}

	\caption{Evaluation on synthetic data, averaged over 10 seeds, as sample size per intervention increases, across different intervention budgets and ground-truth graph densities.
		\textbf{Top}: ARI for evaluating partition recover.
		\textbf{Bottom}: F-score for evaluating edge recovery.
	}
	\label{fig:synth-eval}
\end{figure}

We generate linear Gaussian SCMs on $d=10$ nodes using \texttt{sempler}~\citep{gamella2022characterization} with varying densities (i.e., edge probabilities).
For each SCM, we generate one observational dataset and $\iota \in \{2,5,8\}$ interventional datasets, with per-environment sample size \(n\) ranging from \(10\) to \(10^5\); this is replicated across 10 random seeds.
See \Cref{app:synthetic} for full details as well as a scale-free rather than Erd\H{o}s-R\'enyi version.

\Cref{fig:synth-eval} shows both the ARI (top row) and F-score (bottom row) reliably increasing with $n$ across densities and intervention budgets, empirically corroborating the consistency result in \Cref{cor:asympt-cons}.

For a fixed \(n\) and density, \Cref{fig:synth-eval} shows that the performance decreases with increased intervention budget.
This aligns with theory: more interventions means a finer coarsening, which is harder to learn, with the full DAG being the extreme case.

\begin{figure}[t]
	\centering
	\subfigure[ARI for partition recovery]{
		\includegraphics[width=0.3\textwidth]{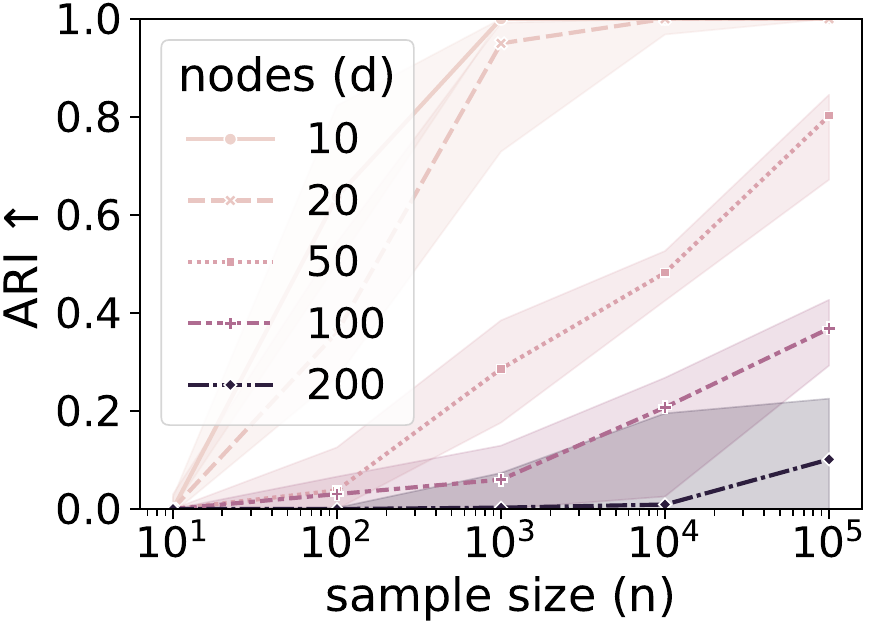}
		\label{fig:scale-ari}
	}
	\hfill
	\subfigure[Run time vs.~sample size]{
		\includegraphics[width=0.3\textwidth]{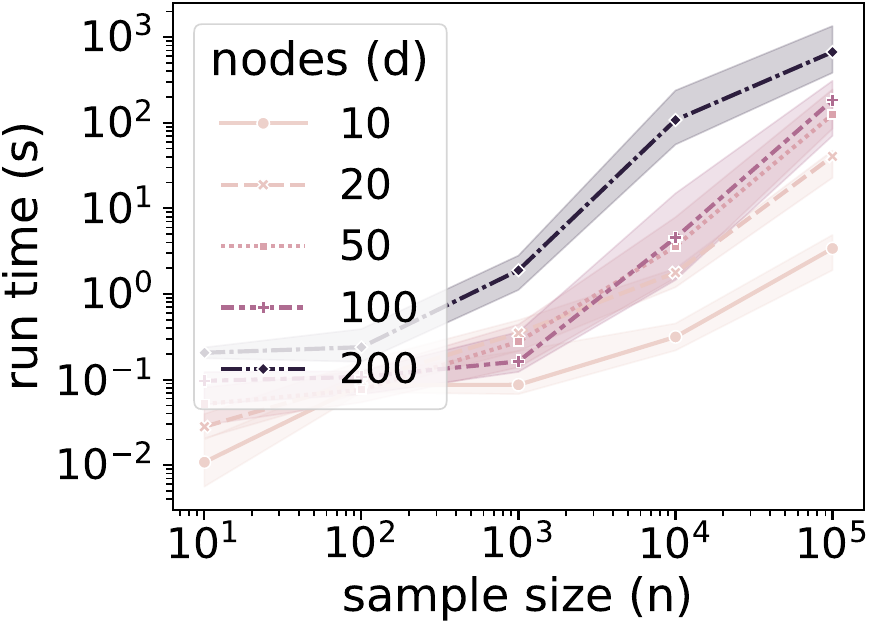}
		\label{fig:scale-t-n}
	}
	\hfill
	\subfigure[Run time vs.~num.~nodes]{
		\includegraphics[width=0.3\textwidth]{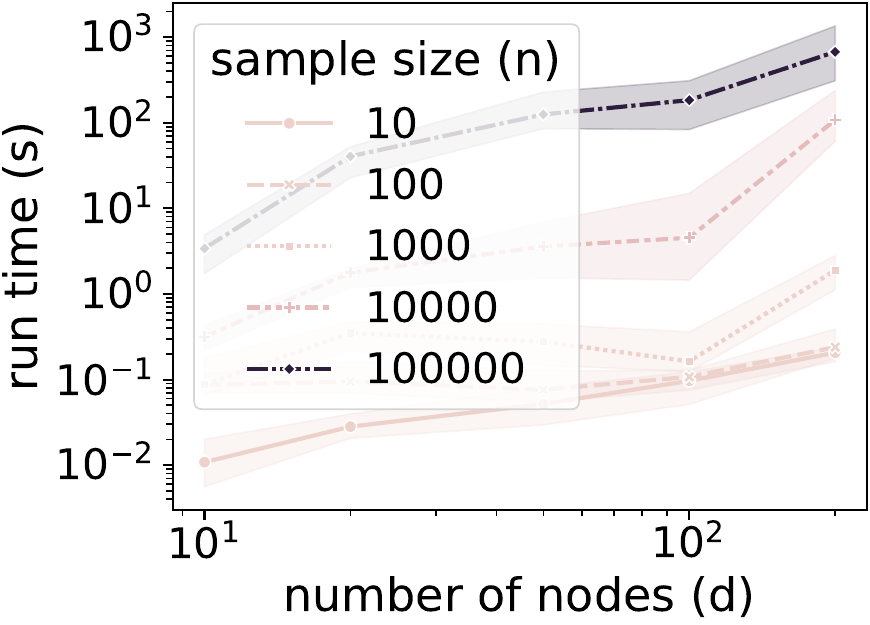}
		\label{fig:scale-t-d}
	}

	\caption{Scalability on synthetic data, averaged over 10 seeds, with density 0.2 and \(\iota=5\).}
	\label{fig:scalability}
\end{figure}

Exploring this further in \Cref{fig:scalability} (see \Cref{app:scalability} for more details), \Cref{fig:scale-ari} demonstrates that, though performance steadily increases with data availability, the performance drops as \(d\) increases for a fixed \(n\)---in other words, larger values of \(d\) require more data to maintain performance, with \(d>50\) requiring \(n>10\,000\).
Nevertheless, the \Cref{fig:scale-t-n,fig:scale-t-d} demonstrate good scalability in terms of run time as \(d\) and \(n\) increase.
Hence, practical limitations in applying \RePaRe stem primarily from statistical power of hypothesis testing and data scarcity rather than the lattice traversal approach itself.

\subsection{Causal Chambers Light Tunnel}
\label{subsec:light-tunnel}

We apply \RePaRe to the \emph{light-tunnel} system from Causal Chambers~\citep{gamellaCausalChambersRealworld2025}, learning an interventional coarsening over a subset of $d=20$ actuator and sensor variables---see \Cref{app:light-tunnel} for full details.
We use interventional data from RGB light intensity perturbations and polarizer-angle perturbations.
We report two variants: \emph{grouped} (pool by intervention type) and \emph{ungrouped} (treat each intervention separately), which ablates intervention granularity.

For hyperparameter selection, \RePaRe has two significance thresholds $(\alpha_{\mathrm{ref}}, \alpha_{\mathrm{edge}})$ used by \RefineTest and \IsEdgeTest.
We perform grid search over these thresholds, selecting via a maximum likelihood estimate (MLE) heuristic score applied to the learned coarsening.
Importantly, this heuristic does not use the ground truth and performs comparably to optimal (given the ground truth) hyperparameter selection.

\begin{sloppypar}
	We compare against three baselines---GIES \citep{hauser2012characterization}, GnIES \citep{gamella2022characterization}, and UT-IGSP \citep{squires2020permutation}---which learn interventional equivalence classes of fine-grained DAGs over the original variables.
	Though the baselines solve a different task than \RePaRe (recovering the refined equivalence class versus learning a coarsening), the results in \Cref{tab:lt_results} are nevertheless informative: \RePaRe achieves close to perfect partition recovery, competitive edge recovery, and orders of magnitude speedups.
\end{sloppypar}

\begin{table}[ht!]
	\centering
	\begin{tabular}{rccccc}
		\toprule
		Method (Selection)           & ARI   & Precision & Recall & F-score & Run time (s) \\
		\midrule
		GIES (score)                 & --    & 0.632     & 0.615  & 0.623   & 7.227        \\
		GnIES (score)                & --    & 0.426     & 0.513  & 0.465   & 1966.916.804 \\
		UT-IGSP (score)              & --    & 0.333     & 0.385  & 0.357   & 1.117        \\
		\midrule
		\RePaRe (grouped, score)     & 0.932 & 1.000     & 0.500  & 0.667   & 0.063        \\
		\RePaRe (grouped, optimal)   & 1.000 & 1.000     & 0.500  & 0.667   & 0.141        \\
		\RePaRe (ungrouped, score)   & 1.000 & 1.000     & 0.800  & 0.889   & 0.753        \\
		\RePaRe (ungrouped, optimal) & 1.000 & 1.000     & 0.800  & 0.889   & 0.753        \\
		\bottomrule
	\end{tabular}
	\caption{Light-tunnel real data results, \texttt{RePaRe} compared to baselines.}
	\label{tab:lt_results}
\end{table}

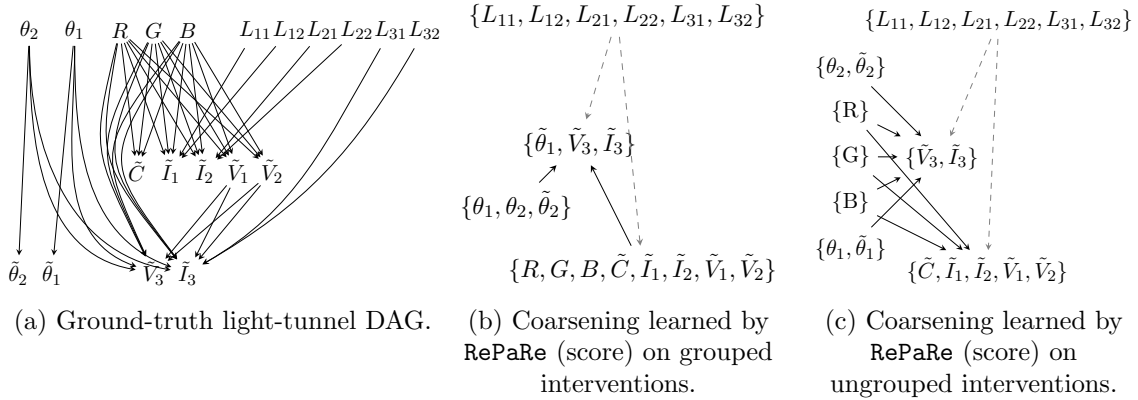
\begin{figure}[t]
	\centering

	\begin{minipage}[t]{0.4\linewidth}
		\centering
		\resizebox{\linewidth}{!}{
			\begin{tikzpicture}[>=stealth, auto=false, node distance=1.2cm]
				\node[circle, inner sep=0.06em] (R)  at (0, 3.7) {$R$};
				\node[circle, inner sep=0.06em] (G)  at (0.6, 3.7) {$G$};
				\node[circle, inner sep=0.06em] (B)  at (1.2, 3.7) {$B$};
				\node[circle, inner sep=0.06em] (T1) at (-0.8, 3.7) {$\theta_1$};
				\node[circle, inner sep=0.06em] (T2) at (-1.6, 3.7) {$\theta_2$};

				\node[circle, inner sep=0.06em] (L11) at (2.4, 3.7) {$L_{11}$};
				\node[circle, inner sep=0.06em] (L12) at (3, 3.7) {$L_{12}$};
				\node[circle, inner sep=0.06em] (L21) at (3.6, 3.7) {$L_{21}$};
				\node[circle, inner sep=0.06em] (L22) at (4.2, 3.7) {$L_{22}$};
				\node[circle, inner sep=0.06em] (L31) at (4.8, 3.7) {$L_{31}$};
				\node[circle, inner sep=0.06em] (L32) at (5.4, 3.7) {$L_{32}$};

				\node[circle, inner sep=0.06em] (C)  at (0.3, 1.2) {$\tilde{C}$};
				\node[circle, inner sep=0.06em] (I1) at (0.9, 1.2) {$\tilde{I}_1$};
				\node[circle, inner sep=0.06em] (I2) at (1.5, 1.2) {$\tilde{I}_2$};
				\node[circle, inner sep=0.06em] (V1) at (2.1, 1.2) {$\tilde{V}_1$};
				\node[circle, inner sep=0.06em] (V2) at (2.7, 1.2) {$\tilde{V}_2$};

				\node[circle, inner sep=0.06em] (I3)  at (1.2, -0.6) {$\tilde{I}_3$};
				\node[circle, inner sep=0.06em] (V3)  at (0.6, -0.6) {$\tilde{V}_3$};
				\node[circle, inner sep=0.06em] (T1t) at (-1.2, -0.6) {$\tilde{\theta}_1$};
				\node[circle, inner sep=0.06em] (T2t) at (-1.8, -0.6) {$\tilde{\theta}_2$};

				\coordinate (auxL) at (-0.3,1.4); 

				\begin{scope}
					\foreach \src in {R,G,B} {
							\draw[->] (\src) -- (C);
							\draw[->] (\src) -- (I1);
							\draw[->] (\src) -- (I2);
							\draw[->] (\src) -- (V1);
							\draw[->] (\src) -- (V2);
						}

					\draw[->] (R) .. controls (-0.5,1.6) .. (V3);
					\draw[->] (R) .. controls (-0.4,1.4) .. (I3);

					\draw[->] (G) .. controls (-0.45,1.6) .. (V3);
					\draw[->] (G) .. controls (-0.35,1.4) .. (I3);

					\draw[->] (B) .. controls (-0.40,1.6) .. (V3);
					\draw[->] (B) .. controls (-0.30,1.4) .. (I3);

					\draw[->] (T1) -- (T1t);
					\draw[->] (T2) -- (T2t);
					\draw[->] (T1) to[out=-90,in=160] (V3);
					\draw[->] (T1) to[out=-90,in=160] (I3);
					\draw[->] (T2) to[out=-90,in=170] (V3);
					\draw[->] (T2) to[out=-90,in=170] (I3);

					\draw[->] (V1) -- (I3);
					\draw[->] (V1) -- (V3);
					\draw[->] (V2) -- (I3);
					\draw[->] (V2) -- (V3);

					\draw[->] (L11) -- (I1);
					\draw[->] (L12) -- (I1);
					\draw[->] (L21) -- (I2);
					\draw[->] (L22) -- (I2);
					\draw[->] (L31) to[out=-120,in=30] (I3);
					\draw[->] (L32) to[out=-120,in=30] (I3);
				\end{scope}
			\end{tikzpicture}
		}%
		\par\vspace{0.5ex}\small (a) Ground-truth light-tunnel DAG.
	\end{minipage}%
	\hfil
	\begin{minipage}[t]{0.29\linewidth}
		\centering
		\resizebox{\linewidth}{!}{%
			\begin{tikzpicture}[>=stealth, auto]
				\node (L)     at (0.6,  1.2)
				{$\{L_{11},L_{12},L_{21},L_{22},L_{31},L_{32}\}$};
				\node (RGB)   at (1, -2.8)
				{$\{R,G,B,\tilde{C} ,\tilde{I}_1,\tilde{I}_2,\tilde{V}_1,\tilde{V}_2\}$};
				\node (Theta) at (-1, -1.8)
				{$\{\theta_1,\theta_2,\tilde{\theta}_2\}$};
				\node (Far)   at (0,  -0.8)
				{$\{\tilde{\theta}_1,\tilde{V}_3,\tilde{I}_3\}$};

				\draw[->] (RGB)   -- (Far);
				\draw[->] (Theta) -- (Far);

				\draw[->, dashed, gray] (L) -- (RGB);
				\draw[->, dashed, gray] (L) -- (Far);
			\end{tikzpicture}%
		}%
		\par\vspace{0.5ex}\small (b) Coarsening learned by \RePaRe (score) on grouped interventions.
	\end{minipage}%
	\hfill
	\begin{minipage}[t]{0.29\linewidth}
		\centering
		\resizebox{\linewidth}{!}{%
			\begin{tikzpicture}[>=stealth, auto]
				\node (Lens)  at (0.6, 2.4)
				{$\{L_{11},L_{12},L_{21},L_{22},L_{31},L_{32}\}$};
				\node (Red)   at (-2,  0.8) {$\{\mathrm{R}\}$};
				\node (Green) at (-2,  0.) {$\{\mathrm{G}\}$};
				\node (Blue)  at (-2, -0.8) {$\{\mathrm{B}\}$};
				\node (Pol1)  at (-2, -1.6) {$\{\theta_1,\tilde\theta_1\}$};
				\node (Pol2)  at (-2,  1.6) {$\{\theta_2,\tilde\theta_2\}$};
				\node (Near)  at (0.4,  -2.0)
				{$\{\tilde{C},\tilde{I}_1,\tilde{I}_2,\tilde{V}_1,\tilde{V}_2\}$};
				\node (Far)   at (-0.4,   0.0) {$\{\tilde{V}_3,\tilde{I}_3\}$};

				\foreach \src in {Red,Green,Blue}
					{
						\draw[->] (\src) -- (Near);
						\draw[->] (\src) -- (Far);
					}
				\draw[->] (Pol1) -- (Far);
				\draw[->] (Pol2) -- (Far);

				\draw[->, dashed, gray] (Lens) -- (Near);
				\draw[->, dashed, gray] (Lens) -- (Far);

			\end{tikzpicture}%
		}%
		\par\vspace{0.5ex}\small (c) Coarsening learned by \RePaRe (score) on ungrouped interventions.
	\end{minipage}

	\caption{\texttt{RePaRe} on the light-tunnel data.
		Panels show (a) the full ground-truth DAG, (b) the coarsened DAG under grouped interventions, and (c) the coarsened DAG under finer (ungrouped) interventions.
		False negatives in \textcolor{gray}{dashed gray}; no false positives.}
	\label{fig:lt}
\end{figure}

\Cref{tab:lt_results} shows near-perfect to perfect partition recovery for all variants.
While the RePaRe (grouped, score) variant achieves a high ARI of 0.932, the other variants achieve an ARI of 1.0, confirming that the interventions largely recover the intended abstraction level.
The ungrouped variants achieve significantly higher edge recall (0.800) and F-scores (0.889) than the grouped variants, reflecting that finer intervention granularity reveals additional causal relations that are otherwise obscured in finite samples.The learned coarsenings in \Cref{fig:lt} remain interpretable and capture the salient causal structure of the ground truth: color perturbations ($R,G,B$) are grouped separately from polarization perturbations ($\theta_1, \theta_2$), while the lenses ($L_{11}, L_{12}, L_{21}, L_{22}, L_{31}, L_{32}$) are shown as independent sources; and the mediators ($\tilde{V}_1, \tilde{V}_2$) of color perturbations are separated from the common downstream effects ($\tilde{V}_3, \tilde{I}_3$) of both color and polarization perturbations.

\section{Discussion and Limitations}
\label{sec:discussion}

We develop a graphical framework for causal abstraction via coarsening:
we characterize when a map from fine- to coarse-grained variables preserves causal structure (\Cref{def:gen-coarse}) and show that the resulting set of coarsenings forms a sublattice of the partition refinement lattice (\Cref{thm:lattice}).
This yields a reusable foundation: any notion of abstraction that enforces the basic coarsening properties of \Cref{def:gen-coarse} corresponds to restricting attention to a principled subspace of this lattice.

\RePaRe is one concrete realization of this view.
We focus on the interventional coarsening $G^{\mathcal{I}}$ (\Cref{def:ivn-coarse}) because it has a clear causal interpretation---clusters are precisely the variables that are indistinguishable in terms of intervened-ancestor sets---and because it enables favorable worst-case complexity for learning the abstract graph (\Cref{thm:ivn-complex}), with large-sample correctness under our assumptions (\Cref{cor:asympt-cons}).
The primary limitation is statistical: two-sample (refinement) and conditional independence (edge recovery) tests require sufficient sample sizes per intervention;
in practice, performance degrades when sample size $n$ does not grow sufficiently with the number of features $d$.
Results also depend on the intervention set (which determines the attainable abstraction resolution) and can be affected by finite-sample error accumulation from repeated testing.

Future work could attempt to address these limitations with score- rather than constraint-based search.
Another direction could be to explore different concrete realizations of our general framework, as hinted at in our essential and marginal coarsening examples.

\paragraph{Acknowledgements}
The work by AM was supported by Novo Nordisk Foundation grant number NNF20OC0062897.
The work of FM was supported by Novo Nordisk Foundation grant number NNF23OC0085356.
PM received funding from the European Research Council (ERC) under the European Union's Horizon 2020 research and innovation programme (grant agreement No.~883818).
Part of this research was performed while AM and PM were visiting the Institute for Mathematical and Statistical Innovation (IMSI), which is supported by the National Science Foundation (Grant No.~DMS-1929348).
The authors thank the organizers of the IMSI long program \textit{Algebraic Statistics and Our Changing World} for helping start the collaboration as well as Benjamin Hollering, Joseph Johnson, Liam Solus, and Gaetano Tedesco for helpful discussions and feedback on earlier drafts.

\bibliography{main}

\clearpage
\begin{center}
	\Large\bfseries Supplementary Material
\end{center}
\appendix

\section{Additional Related Work}
\label[appendix]{app:related-work}

Here we briefly elaborate on connections to related work, focusing on three perspectives:
(i) causal abstraction learning,
(ii) causal discovery methods that recover a DAG or an interventional equivalence class over the original variables, and
(iii) approaches to causal discovery with grouped or multivariate nodes when the grouping is assumed known.
This appendix highlights the key distinction of our setting: we aim to \emph{discover} the grouping (a valid partition) and its induced abstract DAG directly from data.
See \Cref{tab:related-comparison} for a summarized comparison to related work.

\subsection{Extensions in Causal Abstraction Learning}
Building on early foundational work that defines macro-variables through observational and interventional equivalence \citep{chalupka2016multi}, recent literature has significantly expanded the methodological toolkit for learning causal abstractions directly from data.
Several approaches parameterize the abstraction map using deep learning architectures. For instance, \citet{xia2024neural} introduce neural causal abstractions to flexibly learn high-level causal representations, while \citet{bing2026structural} enforce structural causal constraints at the macro level through the use of causal bottleneck models. Taking a different representational approach, \citet{dacunto2025causal} propose learning these abstractions guided by the semantic embedding principle.

Another crucial direction in this space focuses on the robustness, consistency, and alignment of abstractions across different environments or functional forms. \citet{zennaro2023jointly} address the challenge of jointly learning abstractions that remain consistent across multiple interventional distributions. To handle complex distributional mappings between micro and macro levels, optimal transport has been leveraged to align abstractions \citep{felekis2024causal}, an approach that was subsequently extended to ensure distributionally robust causal abstractions \citep{felekis2025distributionally}. Furthermore, recent theoretical work formalizes the necessary alignment between graphical abstractions (operations on the DAG itself) and functional abstractions (operations on the underlying structural equations) \citep{schooltink2025aligning}.

Our framework can be formalized through the lens of Causal Abstraction (CA) \citep{rubensteincausal, BeckersH19}.
In standard CA, an abstraction can be characterized by a tuple $(\tau, \omega)$, where $\tau$ maps low-level states to high-level states and $\omega$ maps low-level interventions to high-level interventions.
In the context of Coarsening Causal Models, the abstraction map $\tau$ is restricted to a surjective variable partition, denoted as $\chi: V \to V'$, which groups fine-grained variables into coarse clusters.
Crucially, rather than learning an arbitrary intervention map $\omega$, our framework enforces a canonical \textit{constructive} intervention: an intervention on a coarse node $C \in V'$ is defined as the simultaneous intervention on the set of constituent micro-variables $\chi^{-1}(C)$.
Therefore, the learning objective simplifies from finding a general pair $(\tau, \omega)$ to identifying a valid partition $\chi$ such that this canonical $\omega$ preserves causal consistency (i.e., ensuring micro-variables within a cluster share interventional ancestors).
This constraint structures the hypothesis space as a lattice, enabling efficient search via the RePaRe algorithm.

\subsection{Causal Discovery over Original Variables}
Most causal discovery algorithms operate at the finest-grained level, finding a DAG or an equivalence class over the original nodes.
Methods such as GES and its interventional variant GIES search over DAGs or Markov equivalence classes in observational and interventional settings \citep{hauser2012characterization} or under unknown interventions like GnIES \citep{gamella2022characterization}.
Permutation-based procedures such as IGSP \citep{wang2017igsp} or under unknown interventions UT-IGSP \citep{squires2020permutation} target interventional Markov equivalence classes.
Other approaches leverage additional structural assumptions: DirectLiNGAM learns a linear non-Gaussian DAG from observational data \citep{directLiNGAM}, and recent information-theoretic methods \citep{infoTopoCD} and diffusion-model approaches \citep{sanchez2023diffan} recover the causal structure via a topological order.
While powerful, these methods typically do not address the case where the target causal relationships live at a coarser level than the observed features.

\begin{table}[t]
	\centering
	\resizebox{\textwidth}{!}{%
		\begin{tabular}{@{}llcccl@{}}
			\toprule
			Reference                                     & Setting   & Partition         & Edges   & Code                                                                            & Key Distinction                                          \\
			\midrule
			\multicolumn{6}{@{}l}{\emph{Structure learning given a known clustering}}                                                                                                                                                            \\[2pt]
			\quad \citet{entner2012causal}                & Obs       & Given             & Learned & ---                                                                             & Causal order via linear non-Gaussianity                  \\
			\quad \citet{wahl2023vector}                  & Obs       & Given             & Learned & \href{https://github.com/JonasChoice/2GVecCI}{Yes}                              & Pairwise causal direction between two groups             \\
			\quad \citet{anand2025r128}                   & Obs       & Given             & Learned & ---                                                                             & Cluster-level discovery in Markovian systems             \\
			\quad \citet{gobler2025nonlinear}             & Obs       & Given             & Learned & \href{https://github.com/boschresearch/gresit}{Yes}                             & Nonlinear additive noise for grouped variables           \\
			\quad \citet{wahl2024foundations}             & Obs       & Given             & ---     & ---                                                                             & Faithfulness transfer from micro to group level          \\
			\midrule
			\multicolumn{6}{@{}l}{\emph{Causal inference given a known cluster DAG}}                                                                                                                                                             \\[2pt]
			\quad \citet{anand2023cdag}                   & Obs       & Given             & Given   & ---                                                                             & Causal effect identification in cluster DAGs             \\
			\midrule
			\multicolumn{6}{@{}l}{\emph{Jointly learning abstractions and low-/high-level transformations}}                                                                                                                                      \\[2pt]
			\quad \citet{chalupka2016multi}$^\dagger$     & Both      & Learned           & Learned & ---                                                                             & Macro-variables via obs./int.\ equivalence               \\
			\quad \citet{zennaro2023jointly}$^\dagger$    & Int       & Partially Learned & Given   & \href{https://github.com/mattdravucz/jointly-learning-causal-abstraction/}{Yes} & Consistency of abstractions across environments          \\
			\quad \citet{xia2024neural}$^\dagger$         & Obs       & Given/Learned     & Learned & \href{https://github.com/CausalAILab/NeuralCausalAbstractions}{Yes}             & Neural parameterization of abstraction maps              \\
			\quad \citet{felekis2024causal}$^\dagger$     & Int       & Learned           & Given   & \href{https://github.com/yfelekis/COTA}{Yes}                                    & Optimal-transport-based abstraction alignment            \\
			\quad
			\citet{massidda2024learning}$^\dagger$        & Obs       & Learned           & Learned & \href{https://github.com/rmassidda/causabs}{Yes}                                & Learning causal abstractions from data                   \\
			\quad
			\citet{felekis2025distributionally}$^\dagger$ & Int       & Learned           & Learned & \href{https://github.com/yfelekis/DiRoCA}{Yes}                                  & Distributionally robust causal abstractions              \\
			\quad \citet{dacunto2025causal}$^\dagger$     & Obs       & Learned           & ---     & \href{https://github.com/SPAICOM/calsep}{Yes}                                   & Semantic-embedding-guided abstractions                   \\
			\quad \citet{bing2026structural}$^\dagger$    & Obs/Given & Learned           & Given   & \href{https://github.com/simonbing/StructuralCausalBottleneckModels}{Yes}       & Structural constraints via causal bottlenecks            \\
			\midrule
			\multicolumn{6}{@{}l}{\emph{Learning clusters that preserve causal effect identifiability}}                                                                                                                                          \\[2pt]
			\quad \citet{tikka2023clustering}             & ---       & Learned           & Given   & \href{https://github.com/santikka/transit_cluster}{Yes}                         & Clustering that preserves effect identifiability         \\
			\midrule
			\multicolumn{6}{@{}l}{\emph{Learning interventional coarsenings (partition-based)}}                                                                                                                                                  \\[2pt]
			\quad \texttt{RePaRe} (ours)                  & Int       & Learned           & Learned & \href{https://github.com/Alex-Markham/repare}{Yes}                              & Lattice structure; consistency and complexity guarantees \\
			\bottomrule
		\end{tabular}
	}
	\caption{Practical comparison of related approaches to coarse-grained causal modeling.
		\emph{Setting}: whether the method uses observational (Obs), interventional (Int), or both types of data.
		\emph{Partition}: whether the variable grouping is assumed given or learned.
		\emph{Edges}: whether the coarse-grained causal structure is assumed given, learned, or not applicable (---).
		\emph{Code}: whether public open-source code is readily available (linked if Yes).
		$^\dagger$These methods learn a general abstraction map, not restricted to a variable partition.}
	\label{tab:related-comparison}
\end{table}

\subsection{Causal Discovery with Grouped Nodes}
A related line of work studies causal discovery \emph{with grouped or multivariate nodes}.
Early work extends linear non-Gaussian identifiability to infer a causal order among \emph{known} groups \citep{entner2012causal}.
Two-group methods then decide direction between \emph{pre-specified} multivariate sets \citep{wahl2023vector}.
More recently, \citet{gobler2025nonlinear} extend nonlinear additive noise models to random vectors and propose a two-step approach that learns a group-level order before selecting a compatible graph.
Cluster DAGs provide a coarse graphical representation over \emph{given} clusters and support effect identification and cluster-level discovery in Markovian systems \citep{anand2023cdag,anand2025r128}.
Complementarily, recent theory examines when faithfulness transfers from micro-level graphs to graphs over groups, revealing subtle failure modes \citep{wahl2024foundations}.
Our setting differs in a key aspect: we do \emph{not} assume the grouping is given, but instead aim to \emph{discover} a valid partition and its induced abstract DAG.

\section{Proofs for Results in the Main Text}
\label[appendix]{app:proofs-results-main}

\begin{proof}[\Cref{lem:imap}]
	Recall that \(\mathcal{M}(G) \subseteq \mathcal{M}(G')\) is equivalent to stating that for any disjoint sets \(A', B', C' \subseteq V'\), if \(A' \perp_{G'} B' \mid C'\), then \(\chi^{-1}(A') \perp_{G} \chi^{-1}(B') \mid \chi^{-1}(C')\).
	Contrapositively, we show that if the pre-images are \(d\)-connected in \(G\), their images are \(d\)-connected in \(G'\).

	Let \(A = \chi^{-1}(A')\), \(B = \chi^{-1}(B')\), and \(C = \chi^{-1}(C')\).
	Suppose \(A \not\perp_G B \mid C\).
	Then there exists an active path \(\tau\) in \(G\) between some \(a \in A\) and \(b \in B\).
	Consider the sequence of nodes \(\tau'\) in \(G'\) formed by applying \(\chi\) to the nodes in \(\tau\), removing consecutive duplicates.
	We verify that \(\tau'\) constitutes an active path in \(G'\) relative to \(C'\):

	\begin{itemize}
		\item For every edge \(u \to v\) in \(\tau\), \Cref{def:gen-coarse} implies that either \(\chi(u) \to \chi(v)\) is an edge in \(G'\) or \(\chi(u) = \chi(v)\).
		      In either case, the nodes lie on \(\tau'\).
		\item If \(v_i\) is a non-collider on \(\tau\), then \(v_i \notin C\).
		      Since \(C = \chi^{-1}(C')\), we have \(\chi(v_i) \notin C'\).
		      Thus, the non-collider condition is satisfied in \(G'\).
		\item If \(v_j\) is a collider on \(\tau\), then \(v_j \in C\) or some descendant \(v_d\) of \(v_j\) is in \(C\).
		      Notice that by \Cref{def:gen-coarse},\(v_d \in \de[G]{v_j}\) implies \(\chi(v_d) \in \de[G']{\chi(v_j)}\).
		      Thus, \(\chi(v_j)\) or one of its descendants in \(G'\) is in \(\chi(C) = C'\), so the collider condition is satisfied in \(G'\).
	\end{itemize}
\end{proof}

\begin{proof}[\Cref{thm:lattice}]
	Let \(\mathcal{L}\) be the set of all valid coarsenings of \(G\).
	We equip \(\mathcal{L}\) with the standard refinement ordering \(\preceq\).

	First, we establish the existence of the meet.
	Consider \(A, B \in \mathcal{L}\) with partitions \(\Pi_A\) and \(\Pi_B\).
	Their meet in the partition lattice, \(\Pi_{A \wedge B}\), consists of the non-empty intersections \(\{ U \cap V \mid U \in \Pi_A, V \in \Pi_B \}\).

	Next, we verify that \(\Pi_{A \wedge B}\) corresponds to a valid coarsening.
	Following \Cref{def:gen-coarse}, let \(\chi: V \to \Pi_{A \wedge B}\) be the surjection mapping each node to its part and \(E' = \{\chi(u) \to \chi(v) \mid u\to v \in E, \chi(u) \neq \chi(v)\}\) be the induced edges.
	Suppose for contradiction that \(E'\) contains a cycle \(W_1 \to W_2 \to \cdots \to W_k \to W_1\).
	For each edge, there exist \(u_i \in W_i, u_{i+1} \in W_{i+1}\) with \(u_i \to u_{i+1} \in E\).
	Writing \(W_i = U_i \cap V_i\) with \(U_i \in \Pi_A, V_i \in \Pi_B\), we have \(U_i \neq U_{i+1}\) or \(V_i \neq V_{i+1}\).
	Whenever \(U_i \neq U_{i+1}\), the edge \(u_i\to u_{i+1}\) induces \(U_i \to U_{i+1}\) in \(A\)'s induced graph.
	Following the cycle, at least one of \(\Pi_A\) or \(\Pi_B\) admits a cycle in its induced edges, contradicting that \(A\) and \(B\) are DAGs.
	Thus \(E'\) is acyclic and \(\Pi_{A \wedge B}\) is a valid coarsening.

	Finally, notice that the coarsest common coarsening is the trivial partition \(\mathbf{1} = \{V\}\), which is always a valid coarsening.
	Thus, \(\mathcal{L}\) is a finite meet-semilattice with a top element, so \citep[Proposition~3.3.1]{stanley2011enumerative} guarantees that it is a lattice.
	Since \(\mathcal{L}\) is a subset of partitions of \(V\) ordered by refinement, it is a sublattice of the partition refinement lattice.
\end{proof}

\begin{proof}[\Cref{thm:completeness}]
	Let \(\underline{G}\) be the underlying DAG and \(G^* = (\Pi^*, E^*)\) be the target valid coarsening.
	We explicitly construct the oracles as follows:
	\begin{itemize}\setlength{\parindent}{1em}
		\item \Refine-oracle:
		      Given a current partition \(\Pi\), if \(\Pi = \Pi^*\), return \(\emptyset\).
		      Otherwise, if \(\Pi^* \prec \Pi\), there must exist a part \(\pi \in \Pi\) that is not a part in \(\Pi^*\).
		      Since \(\Pi^*\) refines \(\Pi\), \(\pi\) is the union of some subset of parts from \(\Pi^*\), say \(\pi = \bigcup_{j=1}^m \rho_j\) where \(\rho_j \in \Pi^*\).
		      We define the split \(\pi_a, \pi_b\) by partitioning this set of sub-parts.
		      Because \(G^*\) is a valid coarsening, it is a DAG.
		      Let \(G^*[\pi]\) denote the subgraph of \(G^*\) induced by the node set \(\{\rho_1, \dots, \rho_m\}\).
		      This induced subgraph must also be acyclic, so it contains at least one source node.
		      Let \(\rho_1\) be such a source node.
		      We set \(\pi_a = \rho_1\) and \(\pi_b = \bigcup_{j=2}^m \rho_j\).

		      We now verify Condition 1 (acyclicity preservation).
		      Suppose for contradiction that there is a cycle between \(\pi_a\) and \(\pi_b\) in \(\underline{G}\).
		      This means there exist \(u \in \pi_a = \rho_1\) and \(v \in \pi_b\) such that \(v \in \an[\underline{G}]{u}\).
		      By \Cref{def:gen-coarse}, since \((v, u)\) lies on a directed path in \(\underline{G}\), the coarsening property ensures that \(\chi(v)\) is an ancestor of \(\chi(u)\) in \(G^*\).
		      However, \(\chi(u) = \rho_1\) and \(\chi(v) \in \{\rho_2, \dots, \rho_m\}\), so there exists an edge from some part in \(\{\rho_2, \dots, \rho_m\}\) to \(\rho_1\) in \(G^*[\pi]\).
		      This contradicts the choice of \(\rho_1\) as a source node in the acyclic graph \(G^*[\pi]\).
		      Therefore, \(\an[\underline{G}]{\pi_a} \cap \pi_b = \emptyset\), satisfying Condition 1.

		\item \IsEdge-oracle:
		      The oracle returns \texttt{True} for parts \(u, v\) if and only if \(\pa[\underline{G}]{v} \cap u \neq \emptyset\).
		      This satisfies Condition 2 (parent consistency) of \Cref{def:oracles} directly.
		      By \Cref{def:gen-coarse}, an edge \(u \to v\) appears in \(G^*\) if and only if there exist \(x \in u, y \in v\) with \(x \in \pa[\underline{G}]{y}\) and \(u \neq v\).
		      Our \IsEdge-oracle reconstructs exactly this: it returns \texttt{True} whenever \(\pa[\underline{G}]{v} \cap u \neq \emptyset\) (for \(u \neq v\)), which is precisely the condition for an edge in the coarsening.
	\end{itemize}
	Initialized with the trivial partition \(\Pi_0 = \{V\}\), the algorithm iteratively applies these oracles.
	By construction, if \(\Pi^* \prec \Pi_t\) and a split occurs, the new partition \(\Pi_{t+1}\) is formed by splitting a part \(\pi\) into unions of parts from \(\Pi^*\), ensuring \(\Pi^* \preceq \Pi_{t+1}\).
	Since the partition becomes strictly finer at each step and the lattice is finite, the algorithm must terminate.
	Termination occurs if and only if the \Refine-oracle returns \(\{\emptyset,\emptyset\}\), which happens precisely when \(\Pi_t = \Pi^*\).
	Finally, the edge set is constructed at each step by the \IsEdge-oracle according to the definition of a valid coarsening, ensuring the final output is exactly \(G^*\).
\end{proof}

\begin{proof}[\Cref{thm:ivn-coarse}]
	We show that there exist \Refine- and \IsEdge-oracles satisfying \Cref{def:oracles} that can be implemented using only the family of interventional distributions \(\{f^I\}_{I\in\mathcal I}\) and \Cref{asm:ivn-coarse}.
	Together with \Cref{thm:completeness}, this implies that \(G^{\mathcal I}\) is identifiable.

	For each node \(v\in V\) and each intervention \(I\in\mathcal I\setminus\emptyset\), interventional soundness (\Cref{asm:ivn-coarse}.3) along with its converse (which follows from \Cref{asm:ivn-coarse}.1) gives
	\[
		v\in \de[G]{I}
		\quad\Longleftrightarrow\quad
		f^I(X_v) \neq f^{\emptyset}(X_v).
	\]
	Thus, from the distributions alone one can decide for every pair \((v,I)\) whether \(v\) is a descendant of \(I\) in \(G\).

	Define the (observable) intervention signature of a node \(v\) as
	\[
		\sigma(v) \coloneqq \{I\in\mathcal I\setminus\{\emptyset\} \mid f^I(X_v)\neq f^{\emptyset}(X_v)\}.
	\]
	This \(\sigma(v)\) is precisely the set of interventions whose targets have a directed path to \(v\).
	Since \(\sigma(v)\) depends only on \(\ian[G]{v}\), we have \(\sigma(v) = \sigma(w)\) whenever \(\ian[G]{v} = \ian[G]{w}\).

	By \Cref{def:ivn-coarse}, the true partition \(\Pi^{\mathcal I}\) is exactly the partition of \(V\) into signature equivalence classes.

	We construct a \Refine-oracle that uses only the signatures \(\sigma(v)\) and satisfies acyclicity preservation in \Cref{def:oracles}.
	Given the current partition \(\Pi\), if all \(v\in\pi\) share the same signature \(\sigma(v)\) for every \(\pi\in\Pi\), the oracle returns \(\pi^*=\emptyset\).
	Otherwise, there exists some \(\pi\in\Pi\) containing nodes with at least two distinct signatures.
	Fix such a \(\pi\) and choose some \(u\in\pi\).
	Define
	\[
		\pi_a \coloneqq \{v\in\pi \mid \sigma(v)=\sigma(u)\},\qquad
		\pi_b \coloneqq \pi\setminus \pi_a,
	\]
	and set \(\pi^*=\pi\).

	By construction, every node in \(\pi_a\) lies in the same part of \(\Pi^{\mathcal I}\) as \(u\), whereas every node in \(\pi_b\) lies in a different part of \(\Pi^{\mathcal I}\).
	In particular, in the true interventional coarsening \(G^{\mathcal I}\) there is no directed cycle between the coarse nodes containing \(\pi_a\) and \(\pi_b\).

	Suppose for contradiction that there exists a directed cycle in \(G\) alternating between nodes in \(\pi_a\) and \(\pi_b\).
	Since \(\Pi^{\mathcal I}\) is a valid coarsening of \(G\) (by \Cref{def:ivn-coarse}), this cycle would project to a directed cycle in \(G^{\mathcal I}\) via the surjection \(\chi\).
	However, \(G^{\mathcal I}\) is a DAG by assumption, a contradiction.
	Therefore, no such cycle exists, which means either \(\an[G]{\pi_a}\cap \pi_b=\emptyset\) or \(\an[G]{\pi_b}\cap \pi_a=\emptyset\).
	Thus the split \((\pi_a,\pi_b)\) satisfies acyclicity preservation as required by \Cref{def:oracles}.
	Hence this construction defines a valid \Refine-oracle which, starting from the trivial partition, refines until it reaches exactly the partition into signature classes, that is, \(\Pi^{\mathcal I}\).

	As the partition is iteratively refined toward \(\Pi^{\mathcal I}\), edges are incrementally identified among parts using the \IsEdge-oracle.
	By \Cref{asm:ivn-coarse}.1--2, the family \(\{f^I\}_{I\in\mathcal I}\) is Markov and faithful with respect to the coarse DAG \(G^{\mathcal I}\).

	When each part \(\pi^*\) is refined into \(\pi_a\) and \(\pi_b\) (corresponding to refining \(G\) into \(G'\)), the oracle constructs edges (following \Cref{ln:is-edge} of \Cref{alg:repare}), leveraging the partial order \(\preceq_{G}\) accumulated during refinement:
	\begin{enumerate}
		\item between refined parts: The refinement process ensures \(\pi_a \preceq_{G'} \pi_b\) (by signature structure).
		      The oracle tests \(X_{\pi_a} \indep X_{\pi_b} \mid X_{\pa[G]{\pi^*}}\) to determine if there is an edge from \(\pi_a\) to \(\pi_b\).
		\item parents of refined parts: For each part \(\pi \in \pa[G]{\pi^*}\) the oracle tests \(X_{\pi} \indep X_{\pi_a} \mid X_{\pa[G]{\pi^*} \setminus \{\pi\}}\).
		      For \(\pi_b\), the oracle additionally conditions on \(\pi_a\) if it was determined to be a parent in item 1 above: \(X_{\pi} \indep X_{\pi_b} \mid X_{(\pa[G]{\pi^*} \setminus \{\pi\}) \cup \{\pi_a\}}\).
		\item children of refined parts: For each part \(\pi \in \ch[G]{\pi^*}\) and each \(\pi_{\mathrm{new}} \in \{\pi_a, \pi_b\}\), let \(\pi_{\mathrm{sib}}\) be the sibling part \(\{\pi_a, \pi_b\} \setminus \{\pi_{\mathrm{new}}\}\). The oracle tests \(X_{\pi_{\mathrm{new}}} \indep X_{\pi} \mid X_{(\pa[G]{\pi} \setminus \{\pi^*\}) \cup \{\pi_{\mathrm{sib}}\}}\).
	\end{enumerate}

	In line with the ordered Markov property, conditioning on any superset of the true parents (excluding the candidate parent) is sufficient for these tests, provided it does not include descendants.
	The conditioning sets in items 1--3 are chosen to satisfy this: they include known predecessors in the partial order (specifically including the sibling part, which guarantees the set contains all true parents of \(G^{\mathcal I}\) other than the candidate) without including descendants.
	By coarse faithfulness, conditional independence among coarse variables correctly identifies edges in \(G'\).
	The algorithm terminates when \(\Pi = \Pi^{\mathcal I}\), at which point all edges have been identified.

	We have exhibited \Refine- and \IsEdge-oracles that depend only on the distributions \(\{f^I\}_{I\in\mathcal I}\) and satisfy the conditions of \Cref{def:oracles}.
	By \Cref{thm:completeness}, running \RePaRe with these oracles terminates and outputs exactly the interventional coarsening \(G^{\mathcal I}\).
	Therefore \(G^{\mathcal I}\) is identifiable under \Cref{asm:ivn-coarse}.
\end{proof}

\begin{proof}[Proof of \Cref{thm:ivn-complex}]
	We break the algorithm into three phases.

	In Phase 1, \RefineAux computes the intervention descendant matrix $M$ via Welch $t$-tests for each $(v, I) \in V \times \mathcal{I}$.
	Each test takes $O(n)$ time, giving total time $O(edn)$.

	In Phase 2, \RefineTest refines the partition by repeatedly splitting parts that contain nodes with different intervention descendant patterns.
	Starting from one part and ending at $k$ parts requires exactly $k-1$ splits.
	At each split, checking the intervention patterns for all nodes takes $O(de)$ time.
	Thus, Phase 2 takes $O(dek) = O(d^2 e)$ (since $k \leq d$).

	In Phase 3, \IsEdgeTest applies CCA-based CI tests to pairs of parts.
	Each test takes $O(p^2 n + p^3)$ time.
	At iteration $i$ (for $i \in \{1, \ldots, k-1\}$), we split a part $\pi^*_i$ with in-degree $|\pa[G]{\pi^*_i}|$ and out-degree $|\ch[G]{\pi^*_i}|$.
	Since the partition has $i$ parts, $\pi^*_i$ has at most $i-1$ parents and at most $i-1$ children, so $r_i := |\pa[G]{\pi^*_i}| + |\ch[G]{\pi^*_i}| \leq 2(i-1) = O(i)$.
	When splitting $\pi^*_i$ into $\pi_a$ and $\pi_b$, the oracle performs:
	one test between the refined parts, for each in-parent two tests, and for each out-child two tests.
	This gives $O(1 + 2r_i) = O(r_i)$ tests per split.
	Summing across all $k-1$ splits:
	\[
		\sum_{i=1}^{k-1} r_i \leq \sum_{i=1}^{k-1} O(i) = O(k^2).
	\]
	Therefore, Phase 3 performs $O(k^2)$ total CI tests, each costing $O(p^2 n + p^3)$, giving Phase 3 time $O(k^2(p^2 n + p^3))$.

	Summing the three phases, we have $O(edn) + O(d^2 e) + O(k^2(p^2 n + p^3))$.
	When $d \leq n$ (the statistical regime), the second term is dominated by the first, and $p^2 n$ dominates $p^3$, leaving \(O\!\left((de + k^2 p^2)n\right)\).
\end{proof}

\section{Additional Theoretical Results}
\label[appendix]{app:theory}

\begin{theorem}
	\label{thm:hamilton}
	A coarsening poset of a DAG \(G=([d], E)\) is a distributive lattice only if the DAG has a directed \((d-1)\)-path.
\end{theorem}
\begin{proof}
	Suppose the longest directed path in \(G\) is \(p = (v_1, v_2, \ldots, v_m)\) with \(m<d\).
	Extend the partial order induced by \(G\) to a total order \(t = (t_1=v_1, t_2, \ldots, t_d=v_m)\).
	Consider a vertex \(v^\ast = t_i \not\in p\), and let \(v^- = \max_{j<i} \{t_j \in p\}\) while \(v^+ = \min_{j>i} \{t_j \in p\}\).
	Then \(\{v, v'\}, \{v, v^\ast\}, \{v', v^\ast\}\) and \(\{v, v', v^\ast\}\) are all valid partitions.
	This sublattice does not satisfy distributivity (it is just the partition refinement lattice on 3 nodes).
\end{proof}

\Cref{thm:hamilton} tells us that DAGs that are sparse (in the sense of not connecting all nodes by at least a single long path, which can perhaps be related more explicitly to average degree~\citep{meyniel1973condition}) do not result in distributive coarsening lattices.
This is a somewhat negative result, because distributive lattices have nice algebraic properties (e.g., a relation to Gr\"obner bases) that would allow use of computational algebraic tools.

\section{Additional Simulation Results}
\label[appendix]{app:results}

See \texttt{src/expt/workflow/scripts/evaluate.py} in the \href{https://github.com/Alex-Markham/repare/blob/v0.2.0/src/expt/workflow/scripts/evaluate.py}{repo} for implementation details of the F-score and ARI evaluation metrics.

\subsection{Synthetic Data}
\label[appendix]{app:synthetic}

\subsubsection{Data Generation Details}
For each sampled ground-truth DAG, intervention targets are selected uniformly at random without replacement.
Interventions are soft, shift operations as implemented in \texttt{sempler} \citep{gamella2022characterization}.

All variables are standardized before learning.

Edge F-scores are computed by comparing the learned coarsening $G=(\Pi,E)$ to the ground-truth DAG \emph{coarsened under the same learned partition} $\Pi$.
Concretely, we apply the surjection induced by $\Pi$ to the ground-truth DAG to get ground-truth directed edges between the estimated parts, and then we compare the estimated edges to these ground-truth edges.

\subsubsection{Scale-Free Experiments}
For the scale-free setting, we use the same linear Gaussian additive noise model (LGANM) framework implemented in \texttt{sempler}~\citep{gamella2022characterization}, but replace the Erdős–Rényi topology by Barabási–Albert (BA) graphs.
We fix $d=10$ and, for each target density, translate it into an attachment parameter $m = \mathrm{round}(\max(\deg/2, 1))$.
We then sample an undirected BA graph with \texttt{barabasi\_albert\_graph} from \texttt{networkx} library and orient its edges according to a random topological ordering to obtain a DAG with a heavy-tailed degree distribution.
Edge weights are drawn uniformly from $[0.5, 2.0]$ with independent random signs, and we instantiate an LGANM by sampling node-wise means from $[-2, 2]$ and noise
variances from $[0.5, 2]$.

For each ground-truth DAG, we generate one observational dataset and $\iota \in \{2,5,8\}$ interventional datasets.
Intervention targets are selected uniformly at random without replacement, and soft interventions that shift the target mean by $2$ with a variance of $1$.
As in the Erdős–Rényi experiments, we vary the per-environment sample size $n \in [10, 10^5]$ and repeat each $(\text{density}, n, \iota)$ configuration over 10 random seeds.
All variables are standardized before learning.

\begin{figure}[h!]
	\centering
	\subfigure[partition recovery, \(\iota=2\)]{
		\includegraphics[width=0.3\textwidth]{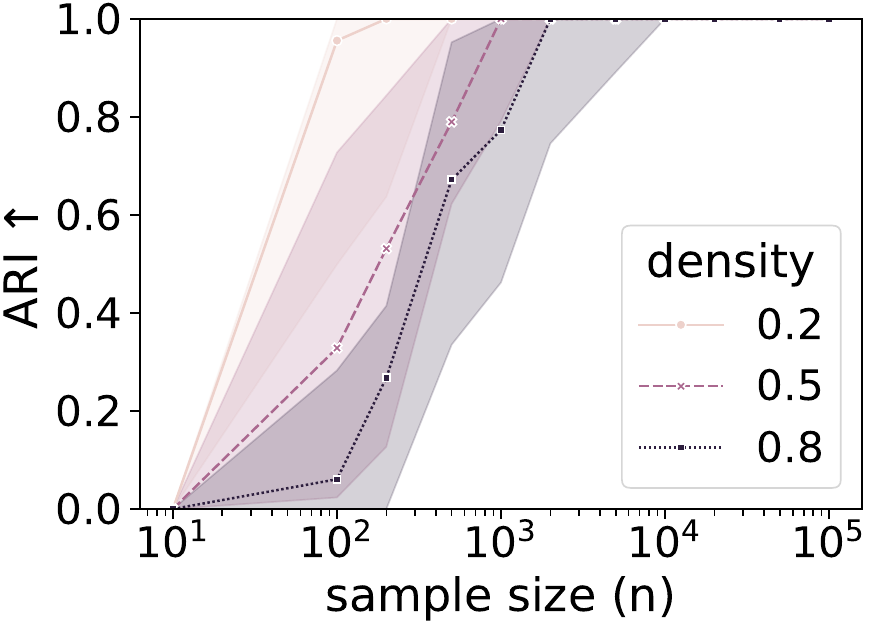}
		\label{fig:ari-k2s}
	}
	\hfill
	\subfigure[partition recovery, \(\iota=5\)]{
		\includegraphics[width=0.3\textwidth]{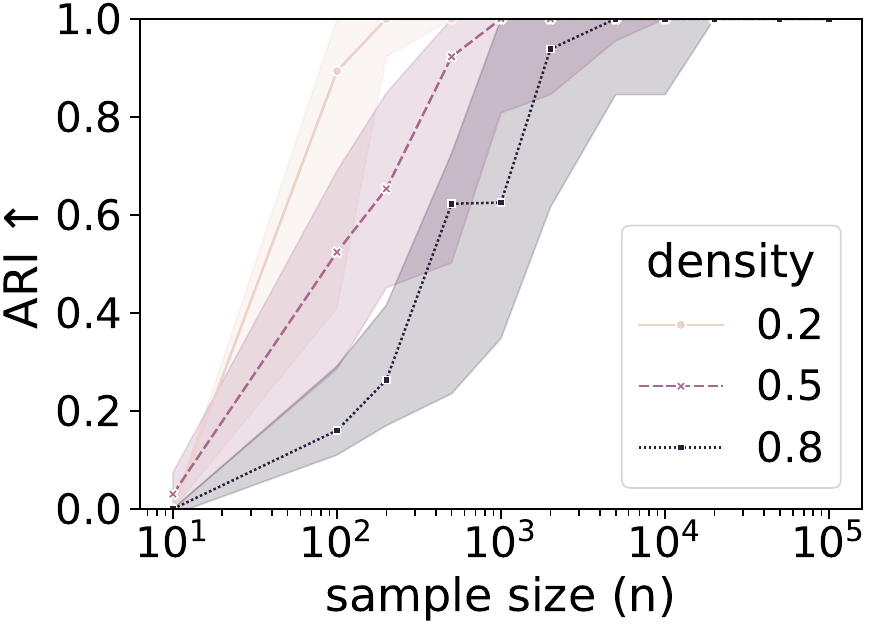}
		\label{fig:ari-k5s}
	}
	\hfill
	\subfigure[partition recovery, \(\iota=8\)]{
		\includegraphics[width=0.3\textwidth]{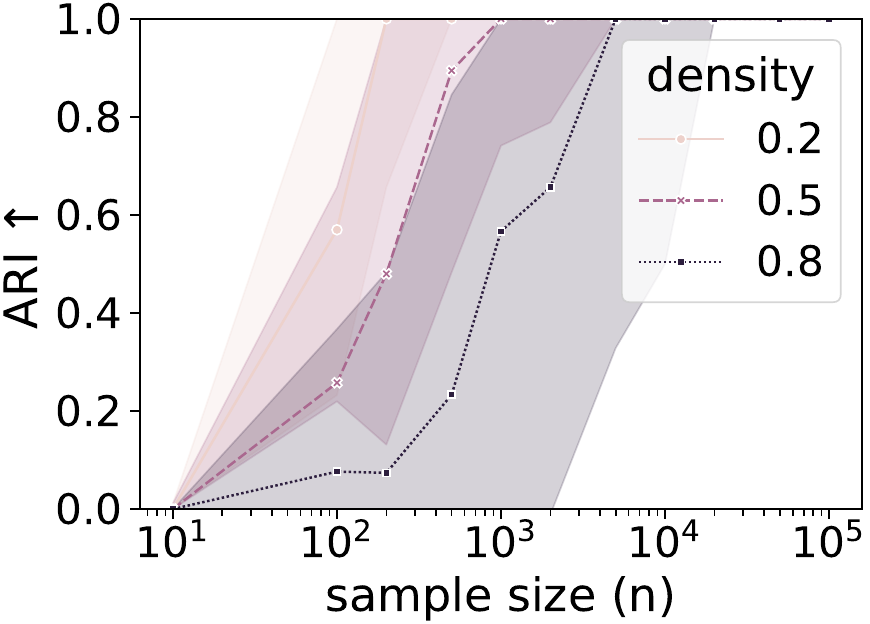}
		\label{fig:ari-k8s}
	}

	\medskip

	\subfigure[edge recovery, \(\iota=2\)]{
		\includegraphics[width=0.3\textwidth]{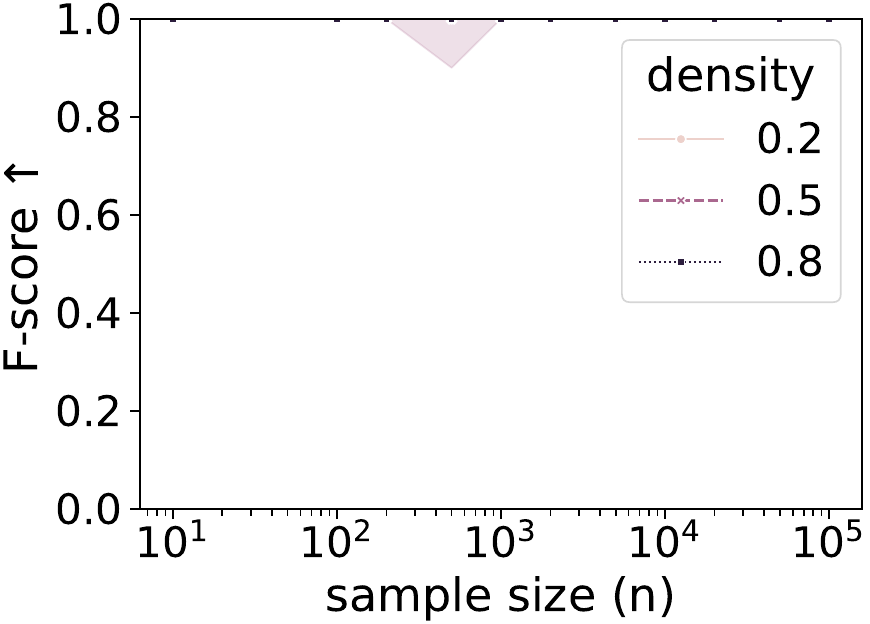}
		\label{fig:fscore-k2s}
	}
	\hfill
	\subfigure[edge recovery, \(\iota=5\)]{
		\includegraphics[width=0.3\textwidth]{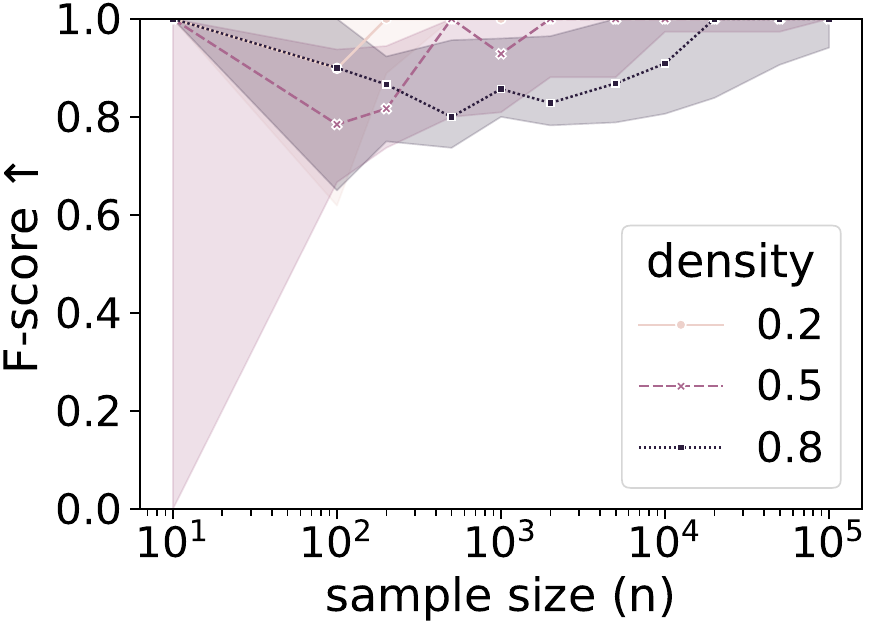}
		\label{fig:fscore-k5s}
	}
	\hfill
	\subfigure[edge recovery, \(\iota=8\)]{
		\includegraphics[width=0.3\textwidth]{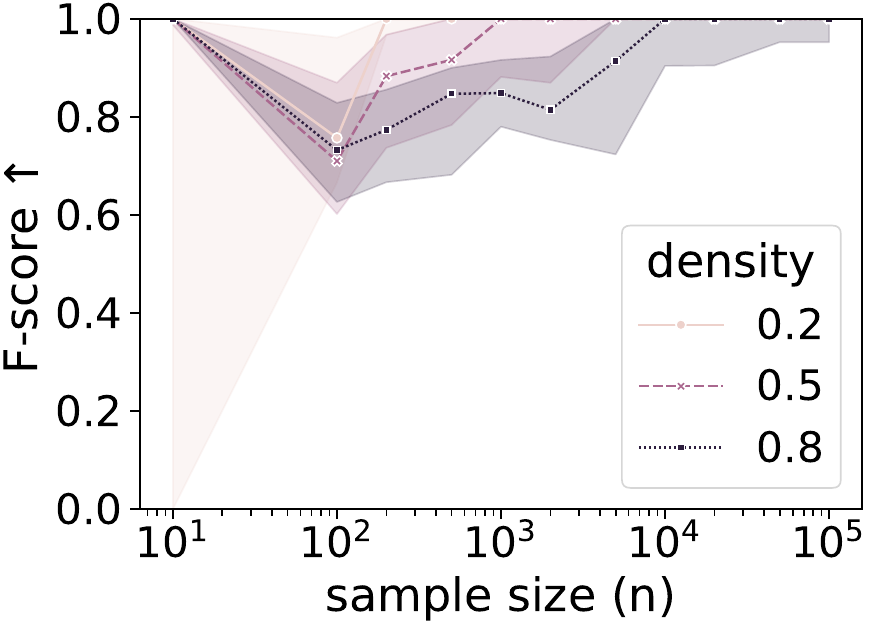}
		\label{fig:fscore-k8s}
	}

	\caption{Evaluation on synthetic data from scale-free DAG models, averaged over 10 seeds, as sample size per intervention increases, across different intervention budgets and ground-truth graph densities.
		\textbf{Top row}: ARI for evaluating partition recover.
		\textbf{Bottom row}: F-score for evaluating edge recovery.
	}
	\label{fig:synth-eval_scale}
\end{figure}

\subsection{Empirical Scalability}
\label[appendix]{app:scalability}

For the empirical computational evaluation, we again use LGANMs, with Erd\H{o}s--R\'enyi.
For each configuration we sample a DAG on $d \in \{10,20,50,100,200\}$ nodes by including each possible edge independently with probability $p=0.2$.
Edge weights are drawn uniformly from $[0.5, 2.0]$ with independent random signs, and we instantiate an LGANM by sampling node-wise means from $[-2, 2]$ and noise variances from $[0.5, 2]$.

For every ground-truth DAG, we generate one observational dataset and $\iota=5$ interventional datasets with single-target interventions.
Targets are selected uniformly at random without replacement, and soft interventions that shift the target mean by $2$ with a variance of $1$, as in the main Erd\H{o}s--R\'enyi experiments.
We vary the per-environment sample size $n \in \{10^2,10^3,10^4,10^5\}$ and repeat each $(d,n)$ configuration over 10 random seeds.
All variables are standardized before learning.

Figure~\ref{fig:scalability} summarizes these results.
For each $d$, the median ARI increases reliably with $n$ but larger graphs require substantially more samples to reach the same level of partition recovery.
This is expected, since the number of possible coarsenings and adjacency relations grows quickly with number of nodes.
At the same time, median runtime grows with both $n$ and $d$, reflecting the fact that \texttt{RePaRe} repeatedly runs multivariate tests over parts whose size and number both increase.
Thus, the regime of good statistical performance is constrained by computational cost in exactly the way suggested by the theory.
Nevertheless, the curves also show that for moderate graph sizes the method is practically usable: for instance, around $d=50$ we already obtain high ARI at sample sizes where wall-clock run time remains well below ten seconds, indicating that \texttt{RePaRe} can reliably recover informative coarsenings on nontrivial graphs within reasonable computational budgets.

\subsection{Light-tunnel Data}
\label[appendix]{app:light-tunnel}

\subsubsection{Light-tunnel variable subset}
\label[appendix]{app:light-tunnel-vars}

We learn a coarsened DAG over the following $d=20$ numeric variables (actuators and sensor readouts):
\begin{align*}
	 & R,\ G,\ B,\ \tilde{C},\
	\tilde{I}_1,\ \tilde{I}_2,\ \tilde{I}_3,\
	\tilde{V}_1,\ \tilde{V}_2,\ \tilde{V}_3, \\
	 & \theta_1,\ \theta_2,\
	\tilde{\theta}_1,\ \tilde{\theta}_2,\
	L_{11},\ L_{12},\
	L_{21},\ L_{22},\
	L_{31},\ L_{32}.
\end{align*}

\subsubsection{Light-tunnel datasets and grouped vs.\ ungrouped construction}
\label[appendix]{app:light-tunnel-details}

We use the observational dataset and five interventional datasets provided for the standard light-tunnel configuration.\footnote{The data and documentation
	are available as the \texttt{lt\_interventions\_standard\_v1} dataset at
	\url{https://github.com/juangamella/causal-chamber}.}
From each experiment we extract a data matrix over the 20 selected variables.

In the grouped setting, we pool runs of the same intervention type, yielding three datasets:
\begin{enumerate}
	\item observational: $X^{\mathrm{obs}}$ from \texttt{uniform\_reference} ($10{,}000$ samples);
	\item RGB pooled: $X^{\mathrm{rgb}}$ by concatenating \texttt{uniform\_red\_strong}, \texttt{uniform\_green\_strong}, \texttt{uniform\_blue\_strong} ($3{,}000$ samples total);
	\item polarizer pooled: $X^{\mathrm{pol}}$ by concatenating \texttt{uniform\_pol\_1\_strong}, \texttt{uniform\_pol\_2\_strong} ($2{,}000$ samples total).
\end{enumerate}
In the ungrouped setting, we treat each of the five intervention experiments as a separate environment rather than pooling by type.
In both cases we assume Gaussianity and instantiate partition refinement via \RefineAux+\RefineTest (based on a \(t\)-test) and adjacency via \IsEdgeTest (Wilks' $\lambda$ CCA test), as in \Cref{sec:learn-coars-caus}.

\subsubsection{Light-tunnel baselines (fine-grained methods)}
\label[appendix]{app:lt-baselines}

All baseline methods estimate directed structure over the 20 fine-grained variables.
\emph{GIES} \citep{hauser2012characterization} is a score-based greedy search method that returns a CPDAG and is in general not consistent \citep{wang2017igsp}.
\emph{UT-IGSP} is a permutation-based procedure that returns a DAG up to the same interventional Markov equivalence class \citep{squires2020permutation}.
Finally, \emph{GnIES} recovers an interventional equivalence class in a linear Gaussian model with unknown intervention targets and returns an interventional essential graph \citep{gamella2022characterization}.

\subsubsection{Model selection heuristic for \texttt{RePaRe}}
\label[appendix]{app:model-selection}

We select hyperparameters without access to the ground truth by ranking candidates using a likelihood heuristic.
For each candidate pair of thresholds $(\alpha_{\mathrm{ref}}, \alpha_{\mathrm{edge}})$, \RePaRe learns an interventional coarsening $G=(\Pi,E)$ using \RefineAux+\RefineTest at level $\alpha_{\mathrm{ref}}$ and \IsEdgeTest at level $\alpha_{\mathrm{edge}}$.

To score this, we form an expansion $\widetilde{G}$ from the paritition back the original variables $V$. In $\widetilde{G}$, we retain all between-part directed edges from $E$. Within each part $\pi \in \Pi$, we assume a fully connected DAG (consistent with a fixed topological ordering \(\leq_\sim\)), which effectively allows for an arbitrary covariance structure within each part.
Formally, \[\widetilde{G} \coloneqq \big(V, \{v \to w \mid \chi(v) \to \chi(w) \in G^{\mathcal{I}} \text{ or } (\chi(v) = \chi(w) \text{ and } v\leq_{\sim} w)\}\big).\]

Let $\mathcal{I}$ be the set of intervention targets (including $\emptyset$ for the observational setting), and let $\mathbf{X}^I \in \mathbb{R}^{n_I \times d}$ be the data matrix for intervention $I \in \mathcal{I}$ with sample size $n_I$. Following \cite{gamella2022characterization}\footnote{We directly use the scoring code in \url{https://github.com/juangamella/gnies} with no penalization term.}, for a weight matrix $B$ compatible with $\widetilde{G}$, we define the precision matrix implied by the graph for intervention $I$ as:$$K^I := [(\text{Id} - B)^{-1}\Omega^I(\text{Id} - B)^{-T}]^{-1},$$where Id denotes the identity matrix. The log-likelihood under a Gaussian model is:$$\log P(\mathbf{X}^I; B, \Omega^I) = -n_I \ln \det(K^I) - n_I \operatorname{tr}(K^I \hat{\Sigma}^I),$$where $\hat{\Sigma}^I$ is the sample covariance for intervention $I$. We rank candidate coarsenings using the maximized interventional likelihood without penalization:

\begin{equation*}
	S(G, \mathcal{I}) = \max_{\substack{B \sim \widetilde{G} \ \\
			\text{diag. pos. def. } {\Omega^I}{I \in \mathcal{I}} \ \\
			\text{s.t. } \mathbb{I}({\Omega^I}{I \in \mathcal{I}}) \subseteq \mathcal{I}}}
	\sum_{I \in \mathcal{I}} \log P(\mathbf{X}^I; B, \Omega^I)
\end{equation*}

Lastly, a coarsening $G=(\Pi,E)$ can be interpreted as a chain graph whose chain components are the parts in $\Pi$.
This connects the heuristic above to Gaussian chain graph MLE theory and intervention semantics \citep{drton2006maximum,lauritzen2002chain}.

\end{document}